\documentclass{article}

\usepackage[preprint]{corl_2026} 
\usepackage{booktabs}
\usepackage{amsmath}    
\usepackage{amssymb}    
\usepackage{amsfonts}   
\usepackage{graphicx}
\usepackage{tikz}
\usetikzlibrary{arrows.meta,positioning,calc}

\usepackage{multirow}

\usepackage{tabularx}
\usepackage{comment}

\usepackage{wrapfig}

\usepackage{enumitem}

\usepackage[table]{xcolor}
\usepackage{xcolor}
\newcommand{\xxnote}[3]{}
\ifx\hidenotes\undefined
  \usepackage{color}
  \renewcommand{\xxnote}[3]{\color{#2}{#1: #3}}
\fi

\title{Robust In-Hand Manipulation via Priors in Reinforcement Learning and Mechanical Design}

\author{
  Yifei Chen\textsuperscript{*} \quad
  Shihan Lu\textsuperscript{*} \quad
  Ed Colgate \quad
  Kevin Lynch\\
  {\normalfont \textsuperscript{*}Equal Contribution} \\
  {\normalfont Center for Robotics and Biosystems, Northwestern University}\\
  \texttt{yifeichen2026@u.northwestern.edu},  
  \texttt{shihanlu@northwestern.edu} \\
}

\begin{document}
\maketitle


\begin{abstract}
In-hand manipulation without external sensing is challenging due to uncertainties from finger-object contacts and disturbances by gravity. While reinforcement learning has shown promise in learning complex finger gaiting, existing approaches do not prioritize maintaining well-conditioned grasps for sustained manipulation. We introduce two complementary physics priors for robust in-hand rolling: a global grasp-quality prior derived from classical grasp analysis and a local contact-geometry prior based on fingertip curvature. The grasp-quality prior is used as a dense reward-shaping term that encourages well-distributed contacts with improved worst-case wrench resistance. The contact-geometry prior is expressed in the fingertip geometry that mechanically shapes the contact interface toward task-aligned rolling while reducing off-axis drift. We evaluate the effect of these priors on learning in-hand rolling manipulation for a multifingered robotic hand manipulating three different objects at four palm orientations. Results show significant improvement in rotation efficiency, grasp stability, and disturbance rejection, suggesting that physics priors embedded in both learning and fingertip morphology improve task robustness and sim-to-real transfer. An overview video can be found at~\url{https://youtu.be/pdd1wHxQnJM?si=dM-U5kiiPTYsk3Pk}.
\end{abstract}



\section{Introduction}

\begin{wrapfigure}{r}{0.42\textwidth}
    \centering
    \vspace{-0.17in}
    \includegraphics[width=0.38\textwidth]{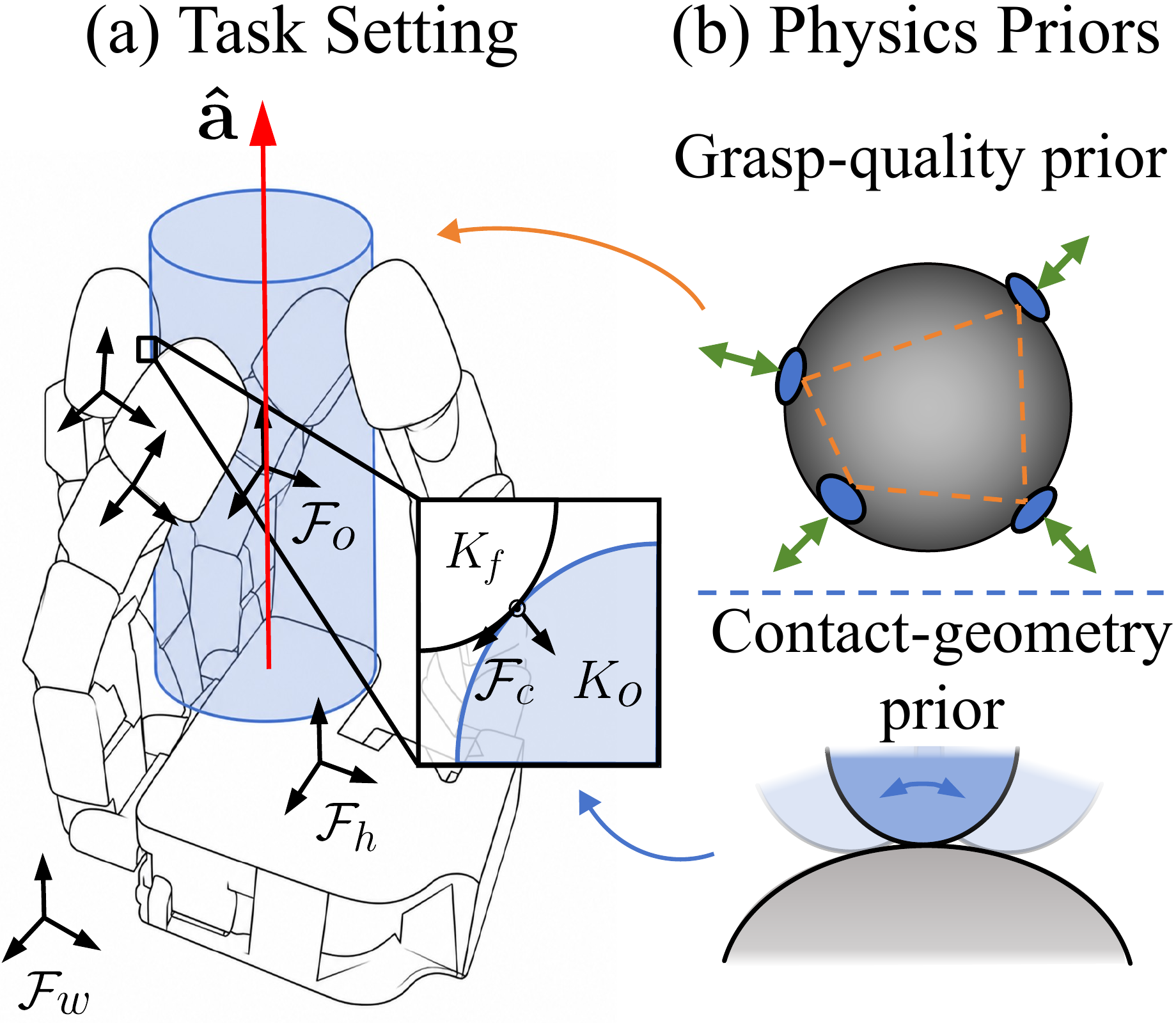}
    \vspace{-0.05in}
    \caption{We embed two physics priors, global grasp quality and local contact geometry, into the learning and fingertip design to improve in-hand rotation.
    }
    \vspace{-0.1in}
    \label{fig:frame_schematic}
\end{wrapfigure}

In-hand manipulation is an essential skill for a wide range of tasks, including regrasping, tool use, and everyday manipulation~\cite{mason1985robot, 6298887, 844067}. This process requires a set of low-level skills that unfold sequentially and concurrently: coordinating multiple fingers to guide object motion, maintaining grasp stability, and regulating finger-object contacts. Each skill is governed by task-related physics priors that directly affect the success and efficiency of manipulation. However, many current approaches  learning in-hand manipulation as a generic policy-learning problem, without explicitly modeling relevant physics underlying these low-level skills~\cite{doi:10.1177/0278364919887447, 10160756, pmlr-v205-qi23a, doi:10.1126/scirobotics.adc9244, doi:10.1126/scirobotics.adl0628}. In this work, we therefore investigate a central question: To what extent can robots leverage task physics to guide and accelerate in-hand manipulation skill learning?

We study continuous in-hand rolling manipulation with a four-finger robot hand. Specifically, the goal is to reorient an object about a fixed axis of the hand through rolling manipulation and finger gaiting under varying palm orientations. We call this \emph{in-hand rotation}. To make the problem more challenging, and to better assess the impact of physics priors on learning in-hand rotation, during execution the robot has access only to proprioceptive feedback, with no tactile or visual feedback. The policy is trained in simulation on randomized cylinders, and evaluated on cylinders, cuboids, and balls in both simulation and the real world to assess generalization beyond the training distribution, using object fall time and rotation speed as performance metrics. 

We study two fundamental and complementary physics priors in in-hand rotation with a multi-fingered hand: (1) a \textit{computationally-imposed} global prior for grasp stability through contact coordination and (2) a \textit{mechanically-imposed} local prior 
based on contact geometry, which can anisotropically alter rolling mobility at a contact. 
We explore how these physics priors facilitate skill acquisition in a reinforcement learning framework and enhance sim-to-real transfer.

For global contact coordination, we inject a physics-based inductive bias based on grasp quality into the learning pipeline, guiding the policy toward stable contact configurations under arbitrary palm orientations, where gravity can substantially alter the manipulation conditions. We further characterize the effect of local fingertip curvature on the stability of in-hand rotation. Our main contributions are as follows:
\begin{itemize} [leftmargin=2em, itemsep=0.1em, topsep=0.15em]
    \item A framework that integrates grasp-quality rewards into a reinforcement learning framework, combined with an anisotropic fingertip design, for robust in-hand rotation. 
    \item A comprehensive evaluation of the impact of the priors in global contact layouts and local contact interfaces on in-hand rotation tasks with a multifingered robot hand, showing that these priors can improve task robustness and sim-to-real transfer. 
\end{itemize}

\textbf{Previous work.} Previous works have studied the problems of multi-fingered in-hand manipulation through both computational algorithmic development~\cite{doi:10.1177/0278364919887447, 10948280, 11128433} and mechanical system design~\cite{bhatt2021surprisingly, 9691924, 10771693}. 
On the algorithmic side, prior work has developed both analytical and learning-based methods to enable finger-gaiting and object reorientation. Analytical methods mathematically formalize contact mechanics, kinematic modeling, and trajectory optimization to generate physically feasible finger motions while satisfying constraints~\cite{677060, 6907059}. In contrast, learning-based methods, primarily reinforcement learning, acquire manipulation policies through interaction and exploration, allowing better finger gaiting under complex dynamics and uncertainty~\cite{10.3389/frobt.2024.1455431, touch-dexterity, bhardwaj2026viserdex}. Despite their differences, both methods face challenges in handling contact transitions and scaling to high-degree-of-freedom hands, requiring either sophisticated modeling or extensive training to ensure robust performance. 
On the mechanical system design end, researchers have explored how hand embodiment~\cite{10.1115/1.4034787, fayhouse}, actuation mechanisms~\cite{9691924, 10771693}, and structural compliance~\cite{bhatt2021surprisingly}, can better handle uncertainty in the physical interactions involved in manipulation. By leveraging the intrinsic properties of hardware, these designs can improve passive stability and contact robustness while optimizing rolling and sliding efficiency, therefore reducing the demands on computation and achieving better reactive behaviors~\cite{7576738, odhner2014compliant}. However, most work in this direction lacks the connection with robot control, often relying on naive or predefined finger gaiting, thus sacrificing general control authority in favor of control simplicity~\cite{7090666, 7473889, 9134855}. An extended related work can be found in Appendix~\ref{app:rextend_related_work}.

Despite recent progress, current robot learning frameworks often lack explicit guarantees of the grasp stability through fast and coordinated contact regulation~\cite{ELGUEAAGUINACO2023102517}. Although tactile sensors are increasingly used in robotic hands, limited temporal bandwidth and sampling rates can make rapid action adjustment difficult under abrupt disturbances~\cite{pmlr-v229-qi23a, doi:10.1126/scirobotics.adl0628, pmlr-v270-yang25c}. Beyond control and sensing, fingertip morphology plays another critical but often overlooked role in in-hand manipulation. Its potential to simplify planning, improve manipulation efficiency, and enhance stability under uncertainty remains largely underexplored, particularly within modern learning frameworks.

\section{Physics-Prior Modeling}
\label{sec:modeling}
 
We assume that in-hand rotation is shaped by two complementary physics priors. At the grasp level, the multi-finger contact configuration should avoid degenerate layouts that are weak against disturbances. At the fingertip contact level, the local fingertip-object interface should permit rolling in the task direction while limiting off-axis motion. 
We model these two properties through a \emph{global grasp-quality prior} derived from the grasp map and a \emph{local contact-geometry prior} based on relative curvature. 
The former is used as a dense learning prior during policy training, while the latter motivates the design of anisotropic fingertip morphology.

\subsection{Task Frames and Model Decomposition}
\label{subsec:modeling_frames}

Let $\mathcal{F}_h$ denote the palm-fixed hand frame and $\mathcal{F}_o$ the object frame at its center of mass, as shown in Fig.~\ref{fig:frame_schematic}. The task is continuous object rotation about a fixed axis in the hand frame, $\hat{\mathbf{a}}\in\mathbb{R}^3$, referred to as the task axis. Defining $\hat{\mathbf{a}}$ in $\mathcal{F}_h$ makes the objective independent of the global palm pose: hand reorientation alters gravity relative to the hand, but not the desired hand-local rotation goal. 

We decompose the model into global grasp-level and local contact-level components. At the grasp level, finger--object contacts, relative to $\mathcal{F}_o$ or $\mathcal{F}_h$, induce a wrench $\mathbf{w}$. This representation forms the basis of our grasp-quality analysis. 

At the contact level, we define a local frame $\mathcal{F}_c=\{X_c,Y_c,Z_c\}$ at each fingertip--object contact, where $Z_c$  aligns with the inward object normal, $X_c$ is in the direction of task-aligned rolling contact motion in the tangent plane, and $Y_c$ completes the right-handed frame. Local object and fingertip curvature matrices $K_o$ and $K_f$ are expressed in this frame.

\subsection{Global Grasp-Quality Prior}
\label{subsec:grasp_quality}

We model continuous in-hand rotation as a sequence of quasistatic finger-object contacts. Motivated by classical grasp analysis~\cite{murray2017mathematical}, we introduce a grasp-quality metric based on the grasp map as an inductive bias in learning, to favor well-distributed contacts and penalize uneven wrench resistance. 

Consider $K$ finger--object contacts relative to the object frame $\mathcal{F}_o$. Let $\mathbf{r}_i \in \mathbb{R}^3$ denote the position of the $i$-th contact point and let $\mathbf{f}_{i} \in \mathbb{R}^3$ denote the corresponding contact force. Stacking all contact forces as $\mathbf{f} = [\mathbf{f}_1^\top,\ldots,\mathbf{f}_K^\top]^\top \in \mathbb{R}^{3K}$, the full object wrench generated by all contacts is
\begin{equation}
    \mathbf{w}
    =
    \begin{bmatrix}
        \mathbf{f}_{\mathrm{obj}} \\
        \boldsymbol{\tau}_{\mathrm{obj}}
    \end{bmatrix}
    =
    \sum_{i=1}^{K}
    \begin{bmatrix}
        \mathbf{I}_3 \\
        [\mathbf{r}_i]_\times \\
    \end{bmatrix}
    \mathbf{f}_{i}
    = 
    \mathbf{G} \mathbf{f},
    \label{eq:reduced_grasp_map}
\end{equation}
where $[\mathbf{r}_i]_\times$ is the skew-symmetric matrix satisfying
$[\mathbf{r}_i]_\times \mathbf{f}_i = \mathbf{r}_i \times \mathbf{f}_i$. 
The matrix $\mathbf{G} = \begin{bmatrix}
    \mathbf{I}_3 \cdots   \mathbf{I}_3 \\
    [\mathbf{r}_1]_\times \cdots [\mathbf{r}_K]_\times 
\end{bmatrix}
\in \mathbb{R}^{6 \times 3K}$ is the grasp map, mapping contact forces to the 6-D object wrench. 

The grasp map $\mathbf{G}$ maps the unit ball in contact-force space to an ellipsoid in the object wrench space, capturing how the contact geometry distributes unit contact forces across different wrench directions.

\begin{figure}[tb]
    \centering
    \includegraphics[width=0.9\linewidth]{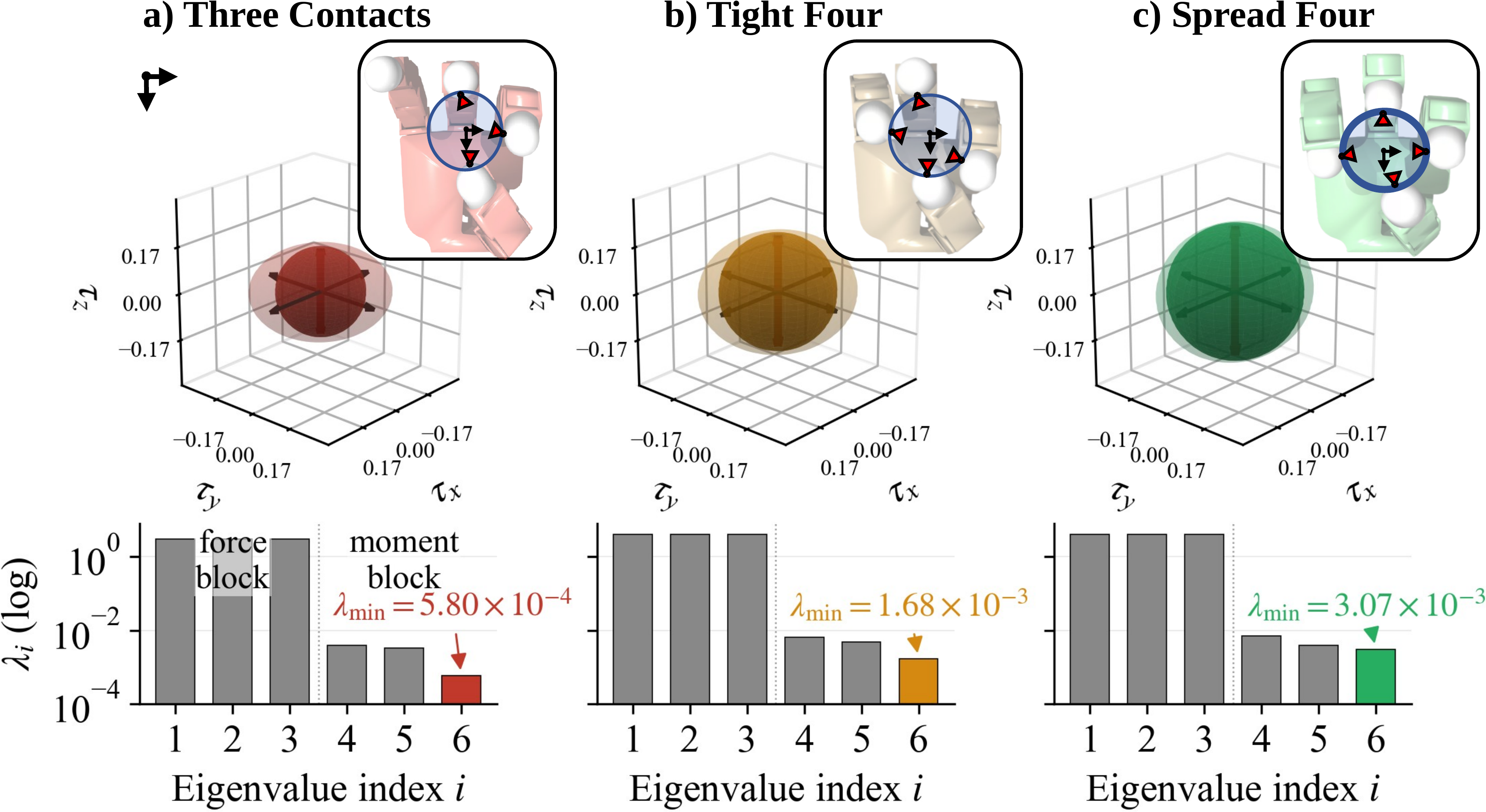}
    \caption{Global grasp quality for three Allegro Hand configurations on a cylinder. 
    Rows show  wrench ellipsoids in the 3-D moment subspace ($\tau_x$, $\tau_y$, $\tau_z$) with the  contact layouts, and Gramian's eigenvalues, with $\lambda_{\min} (\mathbf{M})$ highlighted. More evenly distributed contacts yield larger $\lambda_{\min}$, indicating better worst-case wrench resistance. Visualization details are provided in Appendix~\ref{app:global_grasp_quality}.}
    \label{fig:grasp_quality_global}
\end{figure}

With the simplification of unit contact forces and the wrench-space interpretation of grasp quality following~\cite{li1988task, Roa2015}, we express the contact configuration as the Gramian of the grasp map, 
\begin{equation}
    \mathbf{M} = \mathbf{G}\mathbf{G}^\top .
\end{equation}
The eigenvalues and eigenvectors of $\mathbf{M}\in \mathbb{R}^{6 \times 6}$ determine the principal axes of the corresponding wrench ellipsoid, characterizing the distribution of object wrenches generated by unit contact forces. We define the grasp-quality metric as 
\begin{equation}
  q_{\mathrm{grasp}} = \lambda_{\min} (\mathbf{M}),
  \label{eq:grasp_quality_metric}
\end{equation}
which measures the strength of wrench generation in the weakest direction. Although $q_{\mathrm{grasp}}$ does not  account for friction cones, contact normals, or feasible contact wrench cones as in force closure analysis, it provides a proxy for grasp quality that captures a similar intuition: contact layouts with small minimum eigenvalues of $\mathbf{M}$ are more vulnerable to disturbances along certain wrench directions. We visualize this grasp quality $q_{\mathrm{grasp}}$ along robot hand configurations and wrench ellipsoids in 3D-moment subspace in Fig.~\ref{fig:grasp_quality_global} . 

During training, we use $q_{\mathrm{grasp}}$ as a dense shaping term that biases the policy toward contact distributions with more balanced disturbance resistance. Because it depends only on contact positions, it can be used for policy optimization without requiring force measurements at deployment.

\subsection{Local Contact-Geometry Prior}
\label{subsec:contact_geometry}

Beyond global grasp-quality metric, we further analyze each finger--object contact using local contact kinematics. The analysis focuses on how fingertip geometry can favor rolling along the task direction while discouraging off-axis motion. 

We assume a local rigid-body contact to separate geometry from material compliance and express the contact in the local frame $\mathcal{F}_c=\{X_c,Y_c,Z_c\}$ defined in Sec.~\ref{subsec:modeling_frames}. 
We assume pure rolling, i.e., no sliding at the contact.

Under Montana's contact formulation~\cite{montana1988kinematics}, assuming aligned principal curvature directions with an identity local metric tensor, the fingertip contact-point velocity, $\dot{\mathbf{u}}_f$, reduces to
\begin{equation}
    \dot{\mathbf{u}}_f
    =
    \left(K_o + K_f\right)^{-1}
    \begin{bmatrix}
        -\omega_y \\
        \omega_x
    \end{bmatrix}
    =
    K^{-1}
    \begin{bmatrix}
        -\omega_y \\
        \omega_x
    \end{bmatrix},
    \label{eq:montana_reduced}
\end{equation}
where $K_o$ and $K_f$ are the local object and fingertip curvature matrices, $K = (K_o+K_f)$ is the relative curvature matrix, and $\omega_x$ and $\omega_y$ are the relative rolling angular velocity.

\begin{figure}[b]
    \centering
    \includegraphics[width=\linewidth]{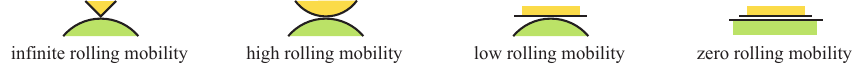}
    \caption{Relative rolling mobility of planar bodies by rolling contact.}
    \label{fig:curvatures}
\end{figure}

As illustrated in Fig.~\ref{fig:curvatures}, high relative curvature corresponds to high rotational mobility by rolling, e.g., for in-hand rolling manipulation. If rolling mobility is high, it may be difficult to precisely control rolling motion, but if rolling mobility is zero, then it is impossible to locally achieve rolling motion between the bodies. Any rotational motion of the object must be achieved by controlled, actuated rotation at the fingertip. 

Assuming a cylindrical object with positive curvature only along $X_c$ and zero curvature along $Y_c$, we set $K_o=\mathrm{diag}(\kappa_o,0)$, where $\kappa_o$ is the curvature in the $X_c$ direction. We compare four fingertip morphologies with nominal curvature $\kappa_f=1/R$: flat, spherical, cylinder-aligned, and cylinder-orthogonal, 
\begin{equation}
    K_f^{\mathrm{flat}}
    =
    \begin{bmatrix}
        0 & 0 \\
        0 & 0
    \end{bmatrix},
    \quad
    K_f^{\mathrm{sphere}}
    =
    \begin{bmatrix}
        \kappa_f & 0 \\
        0 & \kappa_f
    \end{bmatrix}, \quad
    K_f^{\mathrm{aligned}}
    =
    \begin{bmatrix}
        \kappa_f & 0 \\
        0 & 0
    \end{bmatrix},
    \quad
    K_f^{\mathrm{ortho}}
    =
    \begin{bmatrix}
        0 & 0 \\
        0 & \kappa_f
    \end{bmatrix}.
    \label{eq:cyl_curvature}
\end{equation}
Here, the aligned cylindrical fingertip has nonzero curvature along the task direction $X_c$, while the orthogonal variant has nonzero curvature along $Y_c$.

This model reveals the key distinction between isotropic and anisotropic fingertips. A spherical fingertip yields nonzero mobility in both directions, enabling smooth rolling but offering no directional preference. In contrast, the task-aligned cylindrical fingertip keeps the task-aligned mobility nonzero while making the off-axis mobility zero. Thus, the aligned cylindrical morphology induces a rolling direction anisotropy.

In practice, compliance and finite contact patches regularize this idealized model. We view the analysis as a geometric hypothesis: anisotropic fingertip curvature can passively suppress off-axis disturbances while preserving task-aligned rolling, reducing the burden on controllers. We evaluate this hypothesis experimentally.


\section{Methodology}
\label{sec:method}

Sec.~\ref{sec:modeling} identifies two physics priors for in-hand rotation: a grasp-quality metric that detects degenerate contact layouts, and an anisotropic fingertip morphology for promoting task-aligned rolling while reducing off-axis motion. We implement the first as an algorithmic prior in the learning framework, and the second as a hardware prior at the contact interface. 

\subsection{RL Formulation and Policy Training}
\label{subsec:rl_formulation}

\textbf{Formulation.}
We formulate in-hand rotation as a discrete-time Markov decision process (MDP), where the policy outputs a 16-D action that specifies target joint positions for the Allegro Hand, which are tracked by a low-level PD controller. The observation space depends on the training stage, reflecting the information available to the policy in each phase. The policy runs at $20~\mathrm{Hz}$ with a control decimation of six, and each episode lasts $T=400$ policy steps.

\textbf{Training framework.}
\label{subsubsec:rma}
We follow the two-stage Rapid Motor Adaptation (RMA) framework by~\citet{KumarA-RSS-21}, adapted to dexterous manipulation by~\citet{pmlr-v205-qi23a}. RMA trains a privileged teacher policy and distill it into a student policy that infers the relevant latent variables from proprioceptive history. In Stage~I, the teacher receives proprioception and privileged simulation information, including object properties that affect contact dynamics. In Stage~II, the student uses an adaptation module to estimate the privileged latent from a short history of proprioceptive observations. The teacher encoder is frozen, while the adaptation module is trained to match the teacher latent and the policy is fine-tuned with PPO~\cite{schulman2017proximal}. The student policy uses proprioception alone, requires no tactile feedback or object-state measurements, and is transferred zero-shot to the real robot. We highlight two key ingredients that improve training: initial grasp cache and hand-orientation curriculum, with additional architecture and training 
details provided in Appendix~\ref{app:policy_implementation}.

\textbf{Initial grasp cache.} \label{subsubsec:grasp_cache}
To focus training on rotation rather than initial grasp acquisition, we use an offline \emph{initial grasp cache}. Candidate grasps are generated by perturbing a canonical hand pose, closing on the object, and filtering for fingertip proximity within $0.1~\mathrm{m}$, contact by at least three fingertips, valid object height, and no fingertip–palm self-collision. We then rotate each hand–object configuration in $90^{\circ}$ increments about the palm axis and cache only grasps stable across all orientations. Each cached grasp state $\xi =[\mathbf{q}_{h},\mathbf{p}_o,\mathbf{q}_o]\in\mathbb{R}^{23}$ stores a validated initial hand--object configuration, with about $50{,}000$ grasps generated per object scale.

\textbf{Hand-orientation curriculum.}
\label{subsubsec:curriculum} 
We use a four-stage hand-orientation curriculum that progressively expands the reset distribution to expose the policy to more diverse disturbances induced by gravity. At stage $k\in\{0,1,2,3\}$, the maximum perturbation is $\theta_{\max}^{(k)}\in\{0^\circ,45^\circ,90^\circ,180^\circ\}$. At each reset, we sample an axis $\hat{\mathbf{u}}\sim\mathcal{U}(\mathbb{S}^2)$ and angle $\theta\sim\mathcal{U}[-\theta_{\max}^{(k)},\theta_{\max}^{(k)}]$, then rotate the hand–object system by $(\hat{\mathbf{u}},\theta)$. We use this curriculum in both RMA stages. The stage promotion details are in Appendix~\ref{app:policy_implementation}.

\subsection{Reward Design}
\label{subsec:physical_prior}

The reward combines a task term for object rotation, a physics-prior term based on the grasp quality in Sec.~\ref{subsec:grasp_quality}, and regularization terms that discourage grasp degradation and excessive hardware effort. We focus on the physics-prior term here and detail standard remaining terms in Appendix~\ref{app:rewards}. 

\textbf{Grasp-quality shaping.}
Sec.~\ref{subsec:grasp_quality} defines the grasp-quality metric $q_{\mathrm{grasp}}=\lambda_{\min}(\mathbf{M})$, computable from fingertip contact positions alone. We use it as a dense shaping term in a hand-local frame. At each step, we assemble the grasp matrix $\mathbf{G}_h(t)$ from the four fingertip contact positions relative to the object center, expressed in the hand frame via the current hand orientation $\mathbf{q}_h(t)$, and form $\mathbf{M}_h(t)=\mathbf{G}_h(t)\mathbf{G}_h(t)^\top$. The shaping term is
\begin{equation}
    r_{\mathrm{gq}}(t)
    = \beta_{\mathrm{gq}}\,\psi \bigl(\lambda_{\min}(\mathbf{M}_h(t))\bigr),
    \label{eq:fc_reward}
\end{equation}
where $\beta_{\mathrm{gq}}>0$ is the shaping weight and $\psi(\cdot)$ is a bounded monotone function that rewards well-conditioned grasps and penalizes near-degenerate ones; its full form is given in Appendix~\ref{app:rewards}.

Since $\mathbf{G}_h$ is expressed in the hand frame, $r_{\mathrm{gq}}$ is invariant to global palm pose. It depends only on contact positions, requiring no force sensing at deployment. This dense geometric signal aligns with the hand-local task objective and complements the fingertip morphology in Sec.~\ref{subsec:contact_geometry}, which mechanically modulates disturbance sensitivity and rolling mobility at the contact interface.

\subsection{Fingertip Design and Fabrication}
\label{subsec:fingertip}

\begin{wrapfigure}{r}{0.32\textwidth}
    \centering
    \vspace{-0.13in}
    \includegraphics[width=0.32\textwidth]{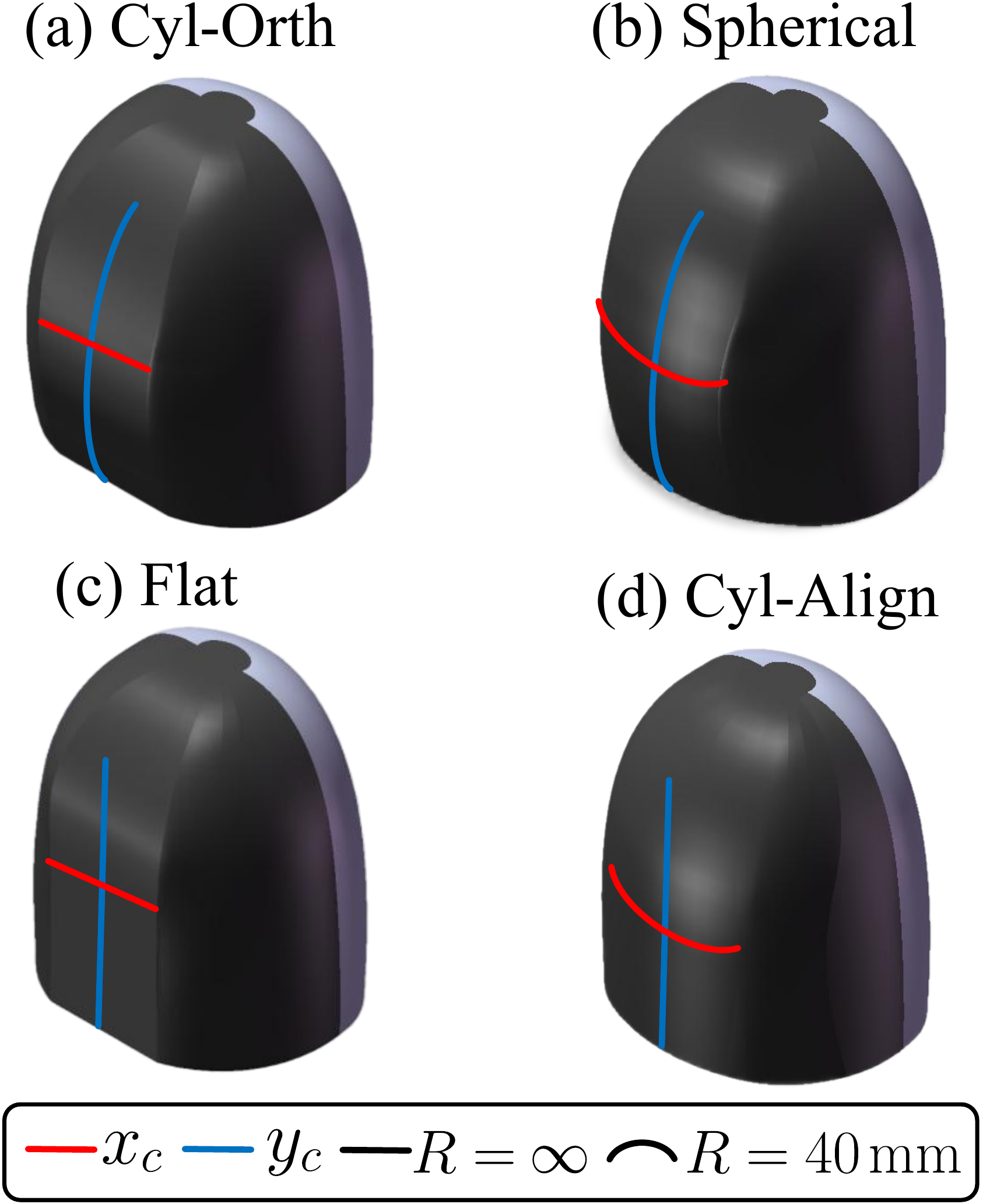}
    \vspace{-0.15in}
    \caption{Candidate fingertip morphologies.}
    \vspace{-0.15in}
    \label{fig:fingertip_design_params}
\end{wrapfigure}

\textbf{Design.}
The four fingertip morphology introduced in
Sec.~\ref{subsec:contact_geometry}---flat, spherical, cylinder-aligned, and cylinder-orthogonal---are parameterized by the principal curvature pair $(\kappa_x,\kappa_y)$ at the nominal contact region. We design all four fingertips' embodiments to share the same overall
envelope, fingertip length, and mounting interface to the Allegro Hand's distal joint, so differences in manipulation behavior are attributed to local surface curvature (Fig.~\ref{fig:fingertip_design_params}). 
The curved variants use radius $R=40~\mathrm{mm}$ ($\kappa=25~\mathrm{m^{-1}}$); the flat variant uses zero curvature in the corresponding direction(s). The fingertip CAD models are imported to the simulation to update the hand model.

\textbf{Fabrication.} 
We fabricate all fingertips by single-piece silicone casting using reusable 3D-printed molds. All molds share the same base, inner insert, and mounting interface. Different morphologies are produced by swapping only the outer cavity halves, ensuring that the fingertips differ only in local contact geometry. We use a two-part platinum-cure silicone rubber (Smooth-On Dragon Skin 30) to preserve a localized contact region while retaining sufficient surface tack for stable rolling. Softer materials can form large contact patches, weakening the point-contact assumption in Sec.~\ref{subsec:contact_geometry}, while harder materials reduce conformity and friction. Fabrication details and outputs are provided in Appendix~\ref{app:fingertip_design}.

\section{Experiments and Evaluation}
\label{sec:experiments}

We evaluate the effects of proposed physics priors on in-hand rotation performance in both simulation and the real world, following our methods in Sec.~\ref{sec:method}. 

\begin{wrapfigure}{r}{0.42\textwidth}
    \centering
    \includegraphics[width=0.42\textwidth]{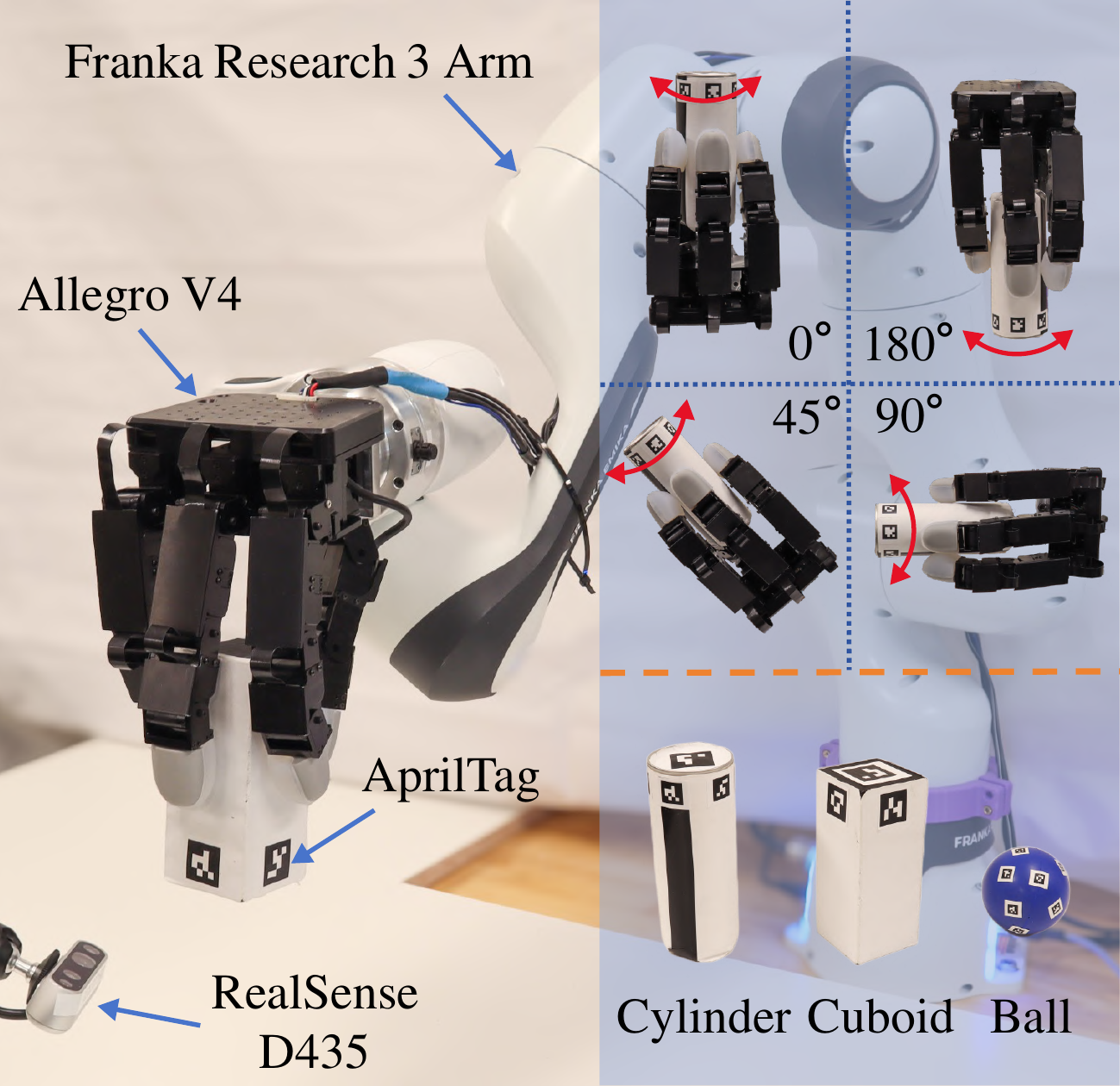}
    \vspace{-0.18in}
    \caption{Experimental setup across four hand orientations and three objects.
    }
    \vspace{-0.25in}
    \label{fig:Exp_setup}
\end{wrapfigure}

\subsection{Setup and Protocol}
\label{subsec:setup}

We use a 16-DoF Allegro V4 four-fingered hand mounted on a Franka Research 3 arm, as the hardware platform. Before each rollout, the arm sets the palm orientation and then remains fixed; the policy does not receive this orientation information. A RealSense D435 camera tracks AprilTags~\cite{5979561} on the object for post-processing only, with no exteroceptive feedback to the policy. In simulation, we use the same Allegro Hand in Isaac Gym~\cite{makoviychuk2021isaac} without modeling the arm.

We train policies in simulation with and without the grasp-quality shaping term for each of the four fingertip designs, and evaluate them in both simulation and hardware. Each condition is trained with three random seeds, and the best policy from each seed is evaluated over three rollouts. Following~\cite{pmlr-v270-yang25c}, we report \textbf{Rot} as mean cumulative rotation about the task axis, \textbf{TTT} as mean episode duration capped at $35~\mathrm{s}$, and \textbf{SR} as gated success rate.

We evaluate three objects, cylinder, cuboid, and sphere, across four palm orientations: $0^\circ,\ 45^\circ,\ 90^\circ$, and $180^\circ$. The objects span distinct surface curvatures to test generalization across contact geometries. The palm reorientation changes the gravity in the hand frame and serves as the primary disturbance. Setup, including object specifications, and details of evaluation and metrics are given in Appendix~\ref{app:experiments}.

\subsection{Evaluation on the Global Grasp-Quality Prior}
\label{subsec:eval_grasp}

We use the default Allegro Hand's fingertip and the aligned cylindrical fingertip as the basis to evaluate the effect of the global grasp-quality (GQ) prior on learning in-hand rotation. Across three objects and four palm tilts, the GQ term improves both rotation efficiency and stability (Table~\ref{tab:cyl_codesign}): overall SR increases from $24\%$ to $56\%$ with the default fingertip (A${\to}$B), and from $64\%$ to $83\%$ with the aligned fingertip (C${\to}$D). TTT also increases from $16.6$ to $22.7\,$s and from $26.1$ to $31.1\,$s, while Rot improves from $0.78$ to $1.25$ and from $1.06$ to $1.27$, respectively.

With the default fingertip, the vanilla reward (A) is fragile across palm tilts, while adding the GQ term (B) improves all metrics, especially at $0^\circ$ and $45^\circ$ tilts. With the aligned fingertip, the GQ term yields the largest gains at larger tilts, $90^\circ$ and $180^\circ$, again improving all three metrics. Combining both priors (D) achieves the best overall SR and TTT by a large margin. The same trend appears in simulation. Simulation results and the real-world per-object results are provided in Appendix~\ref{app:experiments_results}.

\begin{table}[h!]
\centering
\caption{Task performance of the grasp-quality prior on palm orientations across three objects. Gray rows use the default fingertips: row A uses vanilla rewards, and row B adds the grasp-quality reward. White rows use the aligned cylindrical fingertips: row C uses vanilla rewards, and row D adds the grasp-quality reward. The best values in each column are in bold.}
\label{tab:cyl_codesign}
\resizebox{\columnwidth}{!}{
\begin{tabular}{lccccccccccccccc}
\toprule
 & \multicolumn{3}{c}{Tilt 0\textdegree} & \multicolumn{3}{c}{Tilt 45\textdegree} & \multicolumn{3}{c}{Tilt 90\textdegree} & \multicolumn{3}{c}{Tilt 180\textdegree} & \multicolumn{3}{c}{Avg} \\
\cmidrule(lr){2-4} \cmidrule(lr){5-7} \cmidrule(lr){8-10} \cmidrule(lr){11-13} \cmidrule(lr){14-16}
& Rot & TTT & SR & Rot & TTT & SR & Rot & TTT & SR & Rot & TTT & SR & Rot & TTT & SR \\
\midrule
\rowcolor{gray!15}
A
& 0.97 & 17.4 & 30\% & 0.70 & 19.2 & 33\% & 0.43 & 16.7 & 15\% & 1.02 & 13.0 & 19\% & 0.78 & 16.6 & 24\% \\
\rowcolor{gray!15}
B
& \textbf{2.27} & 27.6 & 78\% & 1.04 & 27.0 & 67\% & 0.57 & 19.6 & 37\% & 1.13 & 16.5 & 44\% & 1.25 & 22.7 & 56\% \\
C 
& 1.59 & 33.4 & 81\% & 1.09 & 29.5 & 78\% & 0.59 & 26.5 & 48\% & 0.99 & 15.1 & 48\% & 1.06 & 26.1 & 64\% \\
D
& 1.83 & \textbf{33.6} & \textbf{96\%} & \textbf{1.27} & \textbf{32.9} & \textbf{81\%} & \textbf{0.67} & \textbf{32.5} & \textbf{81\%} & \textbf{1.32} & \textbf{25.4} & \textbf{74\%} & \textbf{1.27} & \textbf{31.1} & \textbf{83\%} \\
\bottomrule
\end{tabular}
}
\end{table}

\subsection{Evaluation on the Local Contact-Geometry Prior}
\label{subsec:eval_morph}

We evaluate the effect of fingertip design across all objects and palm tilts using a fixed policy trained with the grasp-quality term (Table~\ref{tab:morph_gen}). On the cylinder object used in modeling, the aligned cylindrical fingertip is both the most productive and stable design, 
achieving the largest rotation ($2.29$ turns) and longest holding time ($31.9\,$s). The spherical fingertip holds nearly as long 
but rotates much less 
($1.39$ turns), suggesting that it permits rolling without sufficiently constraining motion to the task axis.
The orthogonal fingertip performs worst across all metrics, with the flat fingertip showing comparable results. 
Across objects, the aligned design achieves the highest TTT and SR on all three objects. Its Rot is slightly lower than that of the orthogonal and spherical fingertips 
on the 
ball, whose geometry is less 
aligned with the trained rolling motion. 
Because the simulation has limited contact rendering fidelity, it did not reveal clear trends in the effects of fingertip design.

Breaking down performance by palm orientation with the aligned fingertip design (rows C and D in Table~\ref{tab:cyl_codesign}), the aligned fingertip improves SR at every tilt, with the largest gain at $90^\circ$ tilt ($48\%\!\to\!81\%$) and a large overall improvement across objects ($64\%\!\to\!83\%$). This pattern is consistent with the geometric hypothesis of Sec.~\ref{subsec:contact_geometry}: under $90^\circ$ palm tilts, gravity loads the contacts more laterally on the supporting fingertip located in the bottom, so off-axis ($Y_c$) tangential disturbances become a primary failure mode, causing more severe instability on that finger. The aligned cylindrical fingertip, whose anisotropic curvature is hypothesized to passively suppress motion along $Y_c$, therefore yields its largest gains in this regime.

\begin{table}[h!]
\centering
\caption{Task performance of fingertip designs evaluated on objects across four palm orientations.}
\resizebox{0.9\columnwidth}{!}{
\begin{tabular}{lcccccccccccc}
\toprule
Fingertip & \multicolumn{3}{c}{Cylinder} & \multicolumn{3}{c}{Cuboid} & \multicolumn{3}{c}{Ball} & \multicolumn{3}{c}{Avg} \\
\cmidrule(lr){2-4} \cmidrule(lr){5-7} \cmidrule(lr){8-10} \cmidrule(lr){11-13}
 & Rot & TTT & SR & Rot & TTT & SR & Rot & TTT & SR & Rot & TTT & SR \\
\midrule
Aligned & \textbf{2.29} & \textbf{31.9} & \textbf{86\%} & \textbf{0.69} & \textbf{29.9} & \textbf{83\%} & 0.83 & \textbf{31.6} & \textbf{81\%} & \textbf{1.27} & \textbf{31.1} & \textbf{83\%} \\
Orth    & 0.78 & 21.6 & 67\% & 0.20 & 17.6 & 25\% & \textbf{1.08} & 14.3 & 36\% & 0.68 & 17.8 & 43\% \\
Spherical  & 1.39 & 25.1 & \textbf{86\%} & 0.33 & 20.3 & 39\% & 1.01 & 16.3 & 34\% & 0.91 & 20.6 & 53\% \\
Flat    & 1.46 & 23.3 & 69\% & 0.42 & 26.9 & 50\% & 0.67 & 12.6 & 28\% & 0.85 & 20.9 & 49\% \\
\bottomrule
\end{tabular}
}

\label{tab:morph_gen}
\end{table}

We further isolate the contribution of fingertip geometry by evaluating a single policy, which is trained on the default fingertips, across all four fingertip morphologies with only the physical fingertip swapped at test time. Details and result analysis can be found in Appendix~\ref{app:isolate_fingertip_geometry}.

\section{Conclusion}
\label{sec:conclusion}
 
This work uses in-hand rotation to investigate how robots can leverage task-relevant physics priors to facilitate skill learning. 
We introduce two complementary priors, global grasp quality and local rolling contact, and embed them into the reinforcement learning pipeline and robotic fingertip design. The resulting system achieves in-hand rotation using proprioception alone, while remaining resilient to disturbances induced by varying gravity directions, and significantly improves rotation efficiency, grasp stability, and robustness across objects. These results suggest that explicitly modeling task-relevant physics priors into both learning-based control and mechanical design can provide useful inductive biases for guiding dexterous skill learning and improving sim-to-real transfer.

\section{Limitations}

The generalization of proposed physics priors across broader manipulation categories has not been evaluated. These include in-hand translation, regrasping, and tool-use tasks, which may involve more diverse contact modes, contacts beyond the fingertips, and more complex transitions between rolling, sliding, and sticking. These settings may require additional priors or task-specific biases to capture whole-hand contact interactions and richer object dynamics. Moreover, the claimed robustness against disturbances was primarily demonstrated through changes in gravity direction, without systematic testing under direct external perturbations. Although several demonstration clips suggest that the system can withstand certain external disturbances, a more controlled and quantitative evaluation is needed to substantiate this robustness. In addition, the effectiveness of the proposed method for sim-to-real transfer has not been quantified, making it difficult to determine how much of the real-world performance comes from the proposed priors versus tuning and adaptation during the training and inference.

\clearpage
\acknowledgments{
This material is based upon work primarily supported by the HAND Engineering Research Center (ERC) funded by the ERC Program of the U.S. National Science Foundation under NSF Cooperative Agreement Number EEC-2330040. Any opinions, findings and conclusions, or recommendations expressed in this material are those of the author(s), and do not necessarily reflect those of the NSF.
}


\bibliography{example}  

@ARTICLE{li1988task,
  author={Li, Z. and Sastry, S.S.},
  journal={IEEE Journal on Robotics and Automation}, 
  title={Task-oriented optimal grasping by multifingered robot hands}, 
  year={1988},
  volume={4},
  number={1},
  pages={32-44}}

@article{montana1988kinematics,
  title={The kinematics of contact and grasp},
  author={Montana, David J},
  journal={The International Journal of Robotics Research},
  volume={7},
  number={3},
  pages={17--32},
  year={1988},
  publisher={Sage Publications Sage CA: Thousand Oaks, CA}
}

@inproceedings{Roa2015,
  author    = {M. A. Roa and R. Su{\'a}rez},
  title     = {Grasp Quality Measures: Review and Performance},
  booktitle = {Journal of Autonomous Robots}, 
  year      = {2015},
  pages     = {65--88}
}

@article{doi:10.1177/0278364919887447,
author = {OpenAI: Marcin Andrychowicz and Bowen Baker and Maciek Chociej and Rafal Józefowicz and Bob McGrew and Jakub Pachocki and Arthur Petron and Matthias Plappert and Glenn Powell and Alex Ray and Jonas Schneider and Szymon Sidor and Josh Tobin and Peter Welinder and Lilian Weng and Wojciech Zaremba},
title ={Learning dexterous in-hand manipulation},

journal = {The International Journal of Robotics Research},
volume = {39},
number = {1},
pages = {3-20},
year = {2020}
}

@INPROCEEDINGS{10160756,
  author={Pitz, Johannes and Röstel, Lennart and Sievers, Leon and Bäuml, Berthold},
  booktitle={IEEE International Conference on Robotics and Automation (ICRA)}, 
  title={Dextrous Tactile In-Hand Manipulation Using a Modular Reinforcement Learning Architecture}, 
  year={2023},
  volume={},
  number={},
  pages={1852-1858}}

@InProceedings{pmlr-v205-qi23a,
  title = 	 {In-Hand Object Rotation via Rapid Motor Adaptation},
  author =       {Qi, Haozhi and Kumar, Ashish and Calandra, Roberto and Ma, Yi and Malik, Jitendra},
  booktitle = 	 {Proceedings of The 6th Conference on Robot Learning},
  pages = 	 {1722--1732},
  year = 	 {2023},
  volume = 	 {205}
}

@article{
doi:10.1126/scirobotics.adc9244,
author = {Tao Chen  and Megha Tippur  and Siyang Wu  and Vikash Kumar  and Edward Adelson  and Pulkit Agrawal },
title = {Visual dexterity: In-hand reorientation of novel and complex object shapes},
journal = {Science Robotics},
volume = {8},
number = {84},
pages = {eadc9244},
year = {2023}}

@ARTICLE{10948280,
  author={Yang, Fan and Power, Thomas and Marinovic, Sergio Aguilera and Iba, Soshi and Zarrin, Rana Soltani and Berenson, Dmitry},
  journal={IEEE Robotics and Automation Letters}, 
  title={Multi-Finger Manipulation via Trajectory Optimization With Differentiable Rolling and Geometric Constraints}, 
  year={2025},
  volume={10},
  number={5},
  pages={5170-5177}}

@ARTICLE{9691924,
  author={Morgan, Andrew S. and Hang, Kaiyu and Wen, Bowen and Bekris, Kostas and Dollar, Aaron M.},
  journal={IEEE Robotics and Automation Letters}, 
  title={Complex In-Hand Manipulation Via Compliance-Enabled Finger Gaiting and Multi-Modal Planning}, 
  year={2022},
  volume={7},
  number={2},
  pages={4821-4828}}

@ARTICLE{10771693,
  author={Zhou, Jianshu and Huang, Junda and Dou, Qi and Abbeel, Pieter and Liu, Yunhui},
  journal={IEEE Transactions on Robotics}, 
  title={A Dexterous and Compliant (DexCo) Hand Based on Soft Hydraulic Actuation for Human-Inspired Fine In-Hand Manipulation}, 
  year={2025},
  volume={41},
  number={},
  pages={666-686}}

@inproceedings{bhatt2021surprisingly,
  title={Surprisingly Robust In-Hand Manipulation: An Empirical Study},
  author={Bhatt, Aditya and Sieler, Adrian and Puhlmann, Steffen and Brock, Oliver},
  booktitle={Proceedings of Robotics: Science and Systems},
  year={2021},
}

@inproceedings{fayhouse,
  title={House Of Dextra: Cross-Embodied Co-Design for Dexterous Hands},
  author={Fay, Kehlani and Djapri, Darin Anthony and Zorin, Anya and Clinton, James and El Lahib, Ali and Su, Hao and Tolley, Michael T and Yi, Sha and Wang, Xiaolong},
  booktitle={The Fourteenth International Conference on Learning Representations},
  year={2026}
}

@INPROCEEDINGS{11128433,
  author={Li, Albert H. and Culbertson, Preston and Kurtz, Vince and Ames, Aaron D.},
  booktitle={IEEE International Conference on Robotics and Automation (ICRA)}, 
  title={DROP: Dexterous Reorientation via Online Planning}, 
  year={2025},
  volume={},
  number={},
  pages={14299-14306}}

@ARTICLE{6298887,
  author={Bullock, Ian M. and Ma, Raymond R. and Dollar, Aaron M.},
  journal={IEEE Transactions on Haptics}, 
  title={A Hand-Centric Classification of Human and Robot Dexterous Manipulation}, 
  year={2013},
  volume={6},
  number={2},
  pages={129-144}}

@INPROCEEDINGS{844067,
  author={Okamura, A.M. and Smaby, N. and Cutkosky, M.R.},
  booktitle={IEEE International Conference on Robotics and Automation. Symposia Proceedings (Cat. No.00CH37065}, 
  title={An overview of dexterous manipulation}, 
  year={2000},
  volume={1},
  number={},
  pages={255-262}}

@book{mason1985robot,
  title={Robot hands and the mechanics of manipulation},
  author={Mason, Matthew T and Salisbury Jr, J Kenneth},
  year={1985},
  publisher={MIT press}
}

@INPROCEEDINGS{677060,
  author={Han, L. and Trinkle, J.C.},
  booktitle={IEEE International Conference on Robotics and Automation (Cat. No.98CH36146)}, 
  title={Dextrous manipulation by rolling and finger gaiting}, 
  year={1998},
  volume={1},
  number={},
  pages={730-735}}

@INPROCEEDINGS{6907059,
  author={Bai, Yunfei and Liu, C. Karen},
  booktitle={IEEE International Conference on Robotics and Automation (ICRA)}, 
  title={Dexterous manipulation using both palm and fingers}, 
  year={2014},
  volume={},
  number={},
  pages={1560-1565}}

@InProceedings{pmlr-v229-qi23a,
  title = 	 {General In-hand Object Rotation with Vision and Touch},
  author =       {Qi, Haozhi and Yi, Brent and Suresh, Sudharshan and Lambeta, Mike and Ma, Yi and Calandra, Roberto and Malik, Jitendra},
  booktitle = 	 {Proceedings of The 7th Conference on Robot Learning},
  pages = 	 {2549--2564},
  year = 	 {2023},
  volume = 	 {229}
}

@article{
doi:10.1126/scirobotics.adl0628,
author = {Sudharshan Suresh  and Haozhi Qi  and Tingfan Wu  and Taosha Fan  and Luis Pineda  and Mike Lambeta  and Jitendra Malik  and Mrinal Kalakrishnan  and Roberto Calandra  and Michael Kaess  and Joseph Ortiz  and Mustafa Mukadam },
title = {NeuralFeels with neural fields: Visuotactile perception for in-hand manipulation},
journal = {Science Robotics},
volume = {9},
number = {96},
pages = {eadl0628},
year = {2024}}

@article{bhardwaj2026viserdex,
  title={ViserDex: Visual Sim-to-Real for Robust Dexterous In-hand Reorientation},
  author={Bhardwaj, Arjun and Wilder-Smith, Maximum and Mittal, Mayank and Patil, Vaishakh and Hutter, Marco},
  journal={arXiv preprint arXiv:2604.11138},
  year={2026}
}

@book{murray2017mathematical,
  title={A mathematical introduction to robotic manipulation},
  author={Murray, Richard M and Li, Zexiang and Sastry, S Shankar},
  year={2017},
  publisher={CRC press}
}

@INPROCEEDINGS{5979561,
  author={Olson, Edwin},
  booktitle={IEEE International Conference on Robotics and Automation}, 
  title={AprilTag: A robust and flexible visual fiducial system}, 
  year={2011},
  volume={},
  number={},
  pages={3400-3407}}

@article{makoviychuk2021isaac,
  title={Isaac gym: High performance gpu-based physics simulation for robot learning},
  author={Makoviychuk, Viktor and Wawrzyniak, Lukasz and Guo, Yunrong and Lu, Michelle and Storey, Kier and Macklin, Miles and Hoeller, David and Rudin, Nikita and Allshire, Arthur and Handa, Ankur and others},
  journal={arXiv preprint arXiv:2108.10470},
  year={2021}
}

@INPROCEEDINGS{KumarA-RSS-21, 
    AUTHOR    = {Ashish Kumar AND Zipeng Fu AND Deepak Pathak AND Jitendra Malik}, 
    TITLE     = {{RMA: Rapid Motor Adaptation for Legged Robots}}, 
    BOOKTITLE = {Proceedings of Robotics: Science and Systems}, 
    YEAR      = {2021}
}

@article{schulman2017proximal,
  title={Proximal policy optimization algorithms},
  author={Schulman, John and Wolski, Filip and Dhariwal, Prafulla and Radford, Alec and Klimov, Oleg},
  journal={arXiv preprint arXiv:1707.06347},
  year={2017}
}

@INPROCEEDINGS{219918,
  author={Ferrari, C. and Canny, J.},
  booktitle={IEEE International Conference on Robotics and Automation}, 
  title={Planning optimal grasps}, 
  year={1992},
  volume={},
  number={},
  pages={2290-2295 vol.3}}

@InProceedings{pmlr-v270-yang25c,
  title = 	 {AnyRotate: Gravity-Invariant In-Hand Object Rotation with Sim-to-Real Touch},
  author =       {Yang, Max and lu, chenghua and Church, Alex and Lin, Yijiong and Ford, Christopher J. and Li, Haoran and Psomopoulou, Efi and Barton, David A.W. and Lepora, Nathan F.},
  booktitle = 	 {Proceedings of The 8th Conference on Robot Learning},
  pages = 	 {4727--4747},
  year = 	 {2025},
  volume = 	 {270}
}

@ARTICLE{10.3389/frobt.2024.1455431,
AUTHOR={Weinberg, Abraham Itzhak  and Shirizly, Alon  and Azulay, Osher  and Sintov, Avishai },       
TITLE={Survey of learning-based approaches for robotic in-hand manipulation},       
JOURNAL={Frontiers in Robotics and AI},       
VOLUME={Volume 11},
YEAR={2024}}

@INPROCEEDINGS{7576738,
  author={Or, Keung and Schmitz, Alexander and Funabashi, Satoshi and Tomura, Mami and Sugano, Shigeki},
  booktitle={IEEE International Conference on Advanced Intelligent Mechatronics (AIM)}, 
  title={Development of robotic fingertip morphology for enhanced manipulation stability}, 
  year={2016},
  volume={},
  number={},
  pages={25-30}}

@article{odhner2014compliant,
  title={A compliant, underactuated hand for robust manipulation},
  author={Odhner, Lael U and Jentoft, Leif P and Claffee, Mark R and Corson, Nicholas and Tenzer, Yaroslav and Ma, Raymond R and Buehler, Martin and Kohout, Robert and Howe, Robert D and Dollar, Aaron M},
  journal={The International Journal of Robotics Research},
  volume={33},
  number={5},
  pages={736--752},
  year={2014},
  publisher={SAGE Publications Sage UK: London, England}
}

@INPROCEEDINGS{7090666,
  author={Ma, Raymond R. and Dollar, Aaron M.},
  booktitle={IEEE International Conference on Robotics and Biomimetics (ROBIO)}, 
  title={An underactuated hand for efficient finger-gaiting-based dexterous manipulation}, 
  year={2014},
  volume={},
  number={},
  pages={2214-2219}}

@article{touch-dexterity,
  title          = {Rotating without Seeing: Towards In-hand Dexterity through Touch },
  author         = {Yin, Zhao-Heng and Huang, Binghao and Qin, Yuzhe and Chen, Qifeng and Wang, Xiaolong},
  journal        = {Robotics: Science and Systems},
  year           = {2023},
}

@article{ELGUEAAGUINACO2023102517,
title = {A review on reinforcement learning for contact-rich robotic manipulation tasks},
journal = {Robotics and Computer-Integrated Manufacturing},
volume = {81},
pages = {102517},
year = {2023},
author = {Íñigo Elguea-Aguinaco and Antonio Serrano-Muñoz and Dimitrios Chrysostomou and Ibai Inziarte-Hidalgo and Simon Bøgh and Nestor Arana-Arexolaleiba}
}

@ARTICLE{7473889,
  author={Rojas, Nicolas and Ma, Raymond R. and Dollar, Aaron M.},
  journal={IEEE Transactions on Robotics}, 
  title={The GR2 Gripper: An Underactuated Hand for Open-Loop In-Hand Planar Manipulation}, 
  year={2016},
  volume={32},
  number={3},
  pages={763-770}}

@ARTICLE{9134855,
  author={Abondance, Sylvain and Teeple, Clark B. and Wood, Robert J.},
  journal={IEEE Robotics and Automation Letters}, 
  title={A Dexterous Soft Robotic Hand for Delicate In-Hand Manipulation}, 
  year={2020},
  volume={5},
  number={4},
  pages={5502-5509}}

@inproceedings{ciocarlie2009design,
  title={A design and analysis tool for underactuated compliant hands},
  author={Ciocarlie, Matei and Allen, Peter},
  booktitle={IEEE/RSJ International conference on intelligent robots and systems},
  pages={5234--5239},
  year={2009}
}

@article{aukes2014design,
  title={Design and testing of a selectively compliant underactuated hand},
  author={Aukes, Daniel M and Heyneman, Barrett and Ulmen, John and Stuart, Hannah and Cutkosky, Mark R and Kim, Susan and Garcia, Pablo and Edsinger, Aaron},
  journal={The International Journal of Robotics Research},
  volume={33},
  number={5},
  pages={721--735},
  year={2014},
  publisher={SAGE Publications Sage UK: London, England}
}

@INPROCEEDINGS{Deimel-RSS-14, 
    AUTHOR    = {Raphael Deimel AND Oliver Brock}, 
    TITLE     = {A Novel Type of Compliant, Underactuated Robotic Hand for Dexterous Grasping}, 
    BOOKTITLE = {Proceedings of Robotics: Science and Systems}, 
    YEAR      = {2014}
}

@article{abondance2020dexterous,
  title={A dexterous soft robotic hand for delicate in-hand manipulation},
  author={Abondance, Sylvain and Teeple, Clark B and Wood, Robert J},
  journal={IEEE Robotics and Automation Letters},
  volume={5},
  number={4},
  pages={5502--5509},
  year={2020}
}

@article{
doi:10.1126/sciadv.adu2018,
author = {Ningbin Zhang  and Peiwei Zhou  and Xinyu Yang  and Fengjie Shen  and Jieji Ren  and Tengyu Hou  and Le Dong  and Rong Bian  and Dong Wang  and Guoying Gu  and Xiangyang Zhu },
title = {Biomimetic rigid-soft finger design for highly dexterous and adaptive robotic hands},
journal = {Science Advances},
volume = {11},
number = {17},
pages = {eadu2018},
year = {2025}}

@inproceedings{ciocarlie2010data,
  title={Data-driven optimization for underactuated robotic hands},
  author={Ciocarlie, Matei and Allen, Peter},
  booktitle={IEEE International Conference on Robotics and Automation},
  pages={1292--1299},
  year={2010}
}

@article{puhlmann2022rbo,
  title={RBO hand 3: A platform for soft dexterous manipulation},
  author={Puhlmann, Steffen and Harris, Jason and Brock, Oliver},
  journal={IEEE Transactions on Robotics},
  volume={38},
  number={6},
  pages={3434--3449},
  year={2022}
}

@article{
doi:10.1073/pnas.1003250107,
author = {Eric Brown  and Nicholas Rodenberg  and John Amend  and Annan Mozeika  and Erik Steltz  and Mitchell R. Zakin  and Hod Lipson  and Heinrich M. Jaeger },
title = {Universal robotic gripper based on the jamming of granular material},
journal = {Proceedings of the National Academy of Sciences},
volume = {107},
number = {44},
pages = {18809-18814},
year = {2010}}

@article{PAUL2006619,
title = {Morphological computation: A basis for the analysis of morphology and control requirements},
journal = {Robotics and Autonomous Systems},
volume = {54},
number = {8},
pages = {619-630},
year = {2006},
author = {Chandana Paul}
}

@article{dollar2010highly,
  title={The highly adaptive SDM hand: Design and performance evaluation},
  author={Dollar, Aaron M and Howe, Robert D},
  journal={The international journal of robotics research},
  volume={29},
  number={5},
  pages={585--597},
  year={2010},
  publisher={SAGE Publications Sage UK: London, England}
}

@inproceedings{gupta2016learning,
  title={Learning dexterous manipulation for a soft robotic hand from human demonstrations},
  author={Gupta, Abhishek and Eppner, Clemens and Levine, Sergey and Abbeel, Pieter},
  booktitle={IEEE/RSJ International Conference on Intelligent Robots and Systems (IROS)},
  pages={3786--3793},
  year={2016}
}

@article{10.1115/1.4034787,
    author = {Ma, Raymond R. and Rojas, Nicolas and Dollar, Aaron M.},
    title = {Spherical Hands: Toward Underactuated, In-Hand Manipulation Invariant to Object Size and Grasp Location},
    journal = {Journal of Mechanisms and Robotics},
    volume = {8},
    number = {6},
    pages = {061021},
    year = {2016},
    month = {10}
}

@inproceedings{tincani2012velvet,
  title={Velvet fingers: A dexterous gripper with active surfaces},
  author={Tincani, Vinicio and Catalano, Manuel G and Farnioli, Edoardo and Garabini, Manolo and Grioli, Giorgio and Fantoni, Gualtiero and Bicchi, Antonio},
  booktitle={IEEE/RSJ International Conference on Intelligent Robots and Systems},
  pages={1257--1263},
  year={2012}
}

@article{
doi:10.1126/scirobotics.ado3939,
author = {Jaemin Eom  and Sung Yol Yu  and Woongbae Kim  and Chunghoon Park  and Kristine Yoonseo Lee  and Kyu-Jin Cho },
title = {MOGrip: Gripper for multiobject grasping in pick-and-place tasks using translational movements of fingers},
journal = {Science Robotics},
volume = {9},
number = {97},
pages = {eado3939},
year = {2024}}

@INPROCEEDINGS{9197146,
  author={Yuan, Shenli and Epps, Austin D. and Nowak, Jerome B. and Salisbury, J. Kenneth},
  booktitle={IEEE International Conference on Robotics and Automation (ICRA)}, 
  title={Design of a Roller-Based Dexterous Hand for Object Grasping and Within-Hand Manipulation}, 
  year={2020},
  volume={},
  number={},
  pages={8870-8876}}

@ARTICLE{8239707,
  author={Hawkes, Elliot Wright and Jiang, Hao and Christensen, David L. and Han, Amy K. and Cutkosky, Mark R.},
  journal={IEEE Transactions on Robotics}, 
  title={Grasping Without Squeezing: Design and Modeling of Shear-Activated Grippers}, 
  year={2018},
  volume={34},
  number={2},
  pages={303-316}}

@ARTICLE{10373080,
  author={Xie, Gregory and Holladay, Rachel and Chin, Lillian and Rus, Daniela},
  journal={IEEE Robotics and Automation Letters}, 
  title={In-Hand Manipulation With a Simple Belted Parallel-Jaw Gripper}, 
  year={2024},
  volume={9},
  number={2},
  pages={1334-1341}}

@inproceedings{chen2020hwasp,
	title={Hardware as Policy: Mechanical and ComputationalCo-Optimization using Deep Reinforcement Learning},
	author={Chen, Tianjian and He, Zhanpeng and Ciocarlie, Matei},
	booktitle={Conference on Robotic Learning (CoRL)},
    pages={1158-1173},
	year={2020}
}

@article{doi:10.1177/027836498600400401,
author = {Jeffrey Kerr and Bernard Roth},
title ={Analysis of Multifingered Hands},

journal = {The International Journal of Robotics Research},
volume = {4},
number = {4},
pages = {3-17},
year = {1986}
}

@ARTICLE{28014,
  author={Cole, A.B.A. and Hauser, J.E. and Sastry, S.S.},
  journal={IEEE Transactions on Automatic Control}, 
  title={Kinematics and control of multifingered hands with rolling contact}, 
  year={1989},
  volume={34},
  number={4},
  pages={398-404}}

@INPROCEEDINGS{Sundaralingam-RSS-17, 
    AUTHOR    = {Balakumar Sundaralingam AND Tucker Hermans}, 
    TITLE     = {Relaxed-Rigidity Constraints: In-Grasp Manipulation using Purely Kinematic Trajectory Optimization}, 
    BOOKTITLE = {Proceedings of Robotics: Science and Systems}, 
    YEAR      = {2017}
}

@INPROCEEDINGS{Rajeswaran-RSS-18,

    AUTHOR    = {Aravind Rajeswaran AND Vikash Kumar AND Abhishek Gupta AND

                 Giulia Vezzani AND John Schulman AND Emanuel Todorov AND Sergey Levine},

    TITLE     = "{Learning Complex Dexterous Manipulation with Deep Reinforcement Learning and Demonstrations}",

    BOOKTITLE = {Proceedings of Robotics: Science and Systems (RSS)},

    YEAR      = {2018},

}

@article{HU2025104904,
title = {Dexterous in-hand manipulation of slender cylindrical objects through deep reinforcement learning with tactile sensing},
journal = {Robotics and Autonomous Systems},
volume = {186},
pages = {104904},
year = {2025},
author = {Wenbin Hu and Bidan Huang and Wang Wei Lee and Sicheng Yang and Yu Zheng and Zhibin Li}
}

@INPROCEEDINGS{9340953,
  author={Yuan, Shenli and Shao, Lin and Yako, Connor L. and Gruebele, Alex and Salisbury, J. Kenneth},
  booktitle={IEEE/RSJ International Conference on Intelligent Robots and Systems (IROS)}, 
  title={Design and Control of Roller Grasper V2 for In-Hand Manipulation}, 
  year={2020},
  volume={},
  number={},
  pages={9151-9158}}

@INPROCEEDINGS{Xu-RSS-21, 
    AUTHOR    = {Jie Xu AND Tao Chen AND Lara Zlokapa AND Michael Foshey AND Wojciech Matusik AND Shinjiro Sueda AND Pulkit Agrawal}, 
    TITLE     = {{An End-to-End Differentiable Framework for Contact-Aware Robot Design}}, 
    BOOKTITLE = {Proceedings of Robotics: Science and Systems}, 
    YEAR      = {2021}
}

@article{chen2026ptld,
  title={Ptld: Sim-to-real privileged tactile latent distillation for dexterous manipulation},
  author={Chen, Rosy and Mukadam, Mustafa and Kaess, Michael and Wu, Tingfan and Hogan, Francois R and Malik, Jitendra and Sharma, Akash},
  journal={arXiv preprint arXiv:2603.04531},
  year={2026}
}

@INPROCEEDINGS{11128016,
  author={Qi, Haozhi and Yi, Brent and Lambeta, Mike and Ma, Yi and Calandra, Roberto and Malik, Jitendra},
  booktitle={IEEE International Conference on Robotics and Automation (ICRA)}, 
  title={From Simple to Complex Skills: The Case of In-Hand Object Reorientation}, 
  year={2025},
  volume={},
  number={},
  pages={14291-14298}}

@INPROCEEDINGS{10610532,
  author={Yuan, Ying and Che, Haichuan and Qin, Yuzhe and Huang, Binghao and Yin, Zhao-Heng and Lee, Kang-Won and Wu, Yi and Lim, Soo-Chul and Wang, Xiaolong},
  booktitle={IEEE International Conference on Robotics and Automation (ICRA)}, 
  title={Robot Synesthesia: In-Hand Manipulation with Visuotactile Sensing}, 
  year={2024},
  volume={},
  number={},
  pages={6558-6565}}

@inproceedings{chen2025hand,
  title={In-Hand Manipulation with Enforced Grasp Stability for Contact-Rich Tasks},
  author={Chen, Yifei and Lu, Shihan and Zhang, Haoxuan and Lynch, Kevin M},
  booktitle={ICRA Workshop on Contact-Rich Manipulation},
  year={2025}
}

@INPROCEEDINGS{11127813,
  author={Bauza, Maria and Chen, Jose Enriaue and Dalibard, Valentin and Gileadi, Nimrod and Hafner, Roland and Martins, Murilo F. and Moore, Joss and Pevceviciute, Rugile and Laurens, Antoine and Rao, Dushyant and Zambelli, Martina and Riedmiller, Martin and Scholz, Jon and Bousmalis, Konstantinos and Nori, Francesco and Heess, Nicolas},
  booktitle={IEEE International Conference on Robotics and Automation (ICRA)}, 
  title={DemoStart: Demonstration-Led Auto-Curriculum Applied to Sim-to-Real with Multi-Fingered Robots}, 
  year={2025},
  volume={},
  number={},
  pages={6756-6763}}

\clearpage
\appendix
\section*{Appendix}
\section{Extended Related Work}
\label{app:rextend_related_work}
\subsection{Computational algorithmic approaches}
A substantial body of work has addressed in-hand manipulation through analytical and optimization approaches, using kinematic planning, grasp mechanics, and trajectory or force optimization to generate finger gaiting and reorientation motions under explicitly modeled constraints~\cite{doi:10.1177/027836498600400401, 28014, Sundaralingam-RSS-17, 10948280}. These methods provide important theoretical foundations and offer interpretable solutions for contact-rich manipulation, but often depend on accurate system identification and contact modeling, making them sensitive to uncertainty in the real world and difficult to scale to high-degree-of-freedom (DoF) hands, diverse objects, and long-horizon manipulation.

More recently, model-based and model-free reinforcement learning has been widely explored for dexterous in-hand manipulation. By learning through interaction rather than relying on complete analytical models, these methods have demonstrated increasingly complex behaviors, including object reorientation, finger gaiting, and adaptation to disturbances~\cite{Rajeswaran-RSS-18, touch-dexterity, HU2025104904, chen2025hand, 11127813}. A notable example is the OpenAI system that rotated a Rubik's Cube using a 22-DoF five-finger robotic hand, demonstrating the potential of large-scale model-free reinforcement learning for highly dexterous manipulation~\cite{doi:10.1177/0278364919887447}. Another remarkable series of work is to learn sim-to-real dexterous in-hand rotation policies by using proprioception, vision, and touch to adapt to unknown object properties~\cite{11128016, doi:10.1126/scirobotics.adl0628, 10610532}. 
Despite progress in the real-world deployment, many prior methods depend on additional sensing hardware, such as motion-capture systems or tactile sensors, either to directly estimate object states~\cite{bhardwaj2026viserdex} or to approximate features learned from privileged information available only in simulation~\cite{pmlr-v229-qi23a,chen2026ptld}, for sim-to-real transfer. Furthermore, most prior methods are evaluated in the simplified upward-facing hand orientation, leaving robustness to arbitrary hand orientations and varying gravity conditions comparatively underexplored.

\subsection{Mechanical design approaches} 
Using the inherent mechanical properties of the system to handle the contact uncertainties is another widely explored direction for in-hand manipulation~\cite{ciocarlie2009design, aukes2014design, Deimel-RSS-14, abondance2020dexterous}. Instead of relying solely on high-bandwidth sensing and precise contact regulation, many works exploit compliance~\cite{dollar2010highly}, underactuation~\cite{ciocarlie2010data}, soft materials~\cite{puhlmann2022rbo}, and task-specific morphology~\cite{doi:10.1073/pnas.1003250107, PAUL2006619} to passively adapt to uncertain object geometry and contact conditions. Compliant and underactuated hands can absorb modeling errors and distribute contact forces through mechanical reconfiguration, reducing the burden on planning and control while improving robustness in grasping and manipulation~\cite{ciocarlie2009design, aukes2014design}. Soft robotic hands further leverage material compliance to maintain stable contacts under uncertainty and enable simple in-hand motion primitives~\cite{Deimel-RSS-14, doi:10.1126/sciadv.adu2018, gupta2016learning}. 

Beyond compliance, several designs embed manipulation structure directly into the hand or finger morphology, such as rotary fingertips~\cite{9340953}, spherical hand mechanisms~\cite{10.1115/1.4034787, 9197146}, or wheeled fingers~\cite{tincani2012velvet, doi:10.1126/scirobotics.ado3939, 8239707, 10373080}, to mechanically bias object motion and reduce undesired slip or rolling directions. These approaches demonstrate that embodiment can simplify dexterous in-hand manipulation by trading some general control authority for passive stability, adaptability, and robustness. However, many methods are evaluated with predefined motions or simplified controllers, leaving open how such mechanical priors can be systematically integrated with modern learning-based policies~\cite{chen2020hwasp, Xu-RSS-21}. This motivates our use of fingertip morphology as a local contact prior that complements learning, rather than replacing feedback control or policy optimization.

\section{Modeling Details}

We list all nomenclature across task frames, grasp analysis, and fingertip morphology in Table~\ref{tab:nomenclature} for convenience. 

\begin{table}[htbp]
    \centering
    \begin{minipage}{0.9\textwidth}
    \caption{Nomenclature for frames, contact kinematics, and morphology used throughout the paper.}
    \begin{tabularx}{\linewidth}{@{}lX@{}}
        \toprule
        \textbf{Symbol} & \textbf{Description} \\
        \midrule
        \multicolumn{2}{@{}l}{\textit{Frames}} \\
        $\mathcal{F}_h$ & Hand frame, rigidly attached to the palm \\
        $\mathcal{F}_o$ & Object frame, attached to the object center of mass \\
        $\mathcal{F}_c=\{X_c,Y_c,Z_c\}$ & Local contact frame at a fingertip--object contact \\
        $\hat{\mathbf{a}}$ & Task rotation axis, expressed in $\mathcal{F}_h$ \\
        $X_c$ & Task-aligned tangential axis (desired rolling direction) \\
        $Y_c$ & Orthogonal tangential axis (off-axis disturbance direction) \\
        $Z_c$ & Inward surface normal of the manipulated object \\
        \midrule
        \multicolumn{2}{@{}l}{\textit{Grasp mechanics}} \\
        $K$ & Number of fingertip contacts \\
        $\mathbf{r}_i \in \mathbb{R}^3$ & Position of contact $i$ relative to the object mass centroid \\
        $\mathbf{f}_i \in \mathbb{R}^3$ & Contact force at fingertip $i$ \\
        $\mathbf{w}\in\mathbb{R}^6$ & Object wrench induced by the contacts \\
        $\mathbf{G}\in\mathbb{R}^{6\times 3K}$ & Grasp matrix; $\mathbf{w}=\mathbf{G}\mathbf{f}$ \\
        $\mathbf{M}=\mathbf{G}\mathbf{G}^\top\in\mathbb{R}^{6\times 6}$ & Grasp Gramian \\
        $q_{\mathrm{grasp}}=\lambda_{\min}(\mathbf{M})$ & Grasp-quality metric (minimum eigenvalue of $\mathbf{M}$) \\
        \midrule
         \multicolumn{2}{@{}l}{\textit{Local contact kinematics and morphology}} \\
        $K_o, K_f \in \mathbb{R}^{2\times 2}$ & Local curvature matrices of object and fingertip \\
        $R$ & Nominal fingertip radius \\
        $\kappa_o,\,\kappa_f = 1/R$ & Principal curvatures of object and fingertip surfaces \\
        $\boldsymbol{\omega}_t = [\omega_x,\omega_y]^\top$ & Relative rolling angular velocity in tangent plane \\
        $s_f$ & Contact arc length traversed on the fingertip \\
        $\theta$ & Task-aligned rolling angle about $Y_c$ \\
        \bottomrule
    \end{tabularx}

    \label{tab:nomenclature}
    \end{minipage}
\end{table}

\subsection{Details of the Global Grasp-Quality Prior }
\label{app:global_grasp_quality}
In Fig.~\ref{fig:grasp_quality_global}, we construct three representative grasp configurations on a cylinder ($R=40$~mm, $H=80$~mm) directly in joint space, using the Allegro Hand's canonical pose as the geometric reference.

For four-contact configurations, the grasp matrix is $\mathbf{G}=\begin{bmatrix}\mathbf{I}_3 & \cdots & \mathbf{I}_3\\ [\mathbf{r}_1]_\times & \cdots & [\mathbf{r}_4]_\times\end{bmatrix}\in\mathbb{R}^{6\times12}$; 
the grasp Gramian matrix is $\mathbf{M}=\mathbf{G}\mathbf{G}^\top\in\mathbb{R}^{6\times6}$ and the grasp-quality metric is $q_{\text{grasp}}=\lambda_{\min}(\mathbf{M})$. 

For the three-contact configuration, $\mathbf{G}\in\mathbb{R}^{6\times9}$ is built over the contacting fingertips only.

\textbf{Three configurations (columns).} 
    
\textit{(a) 3-finger contact}: a degenerate grasp in which the thumb has disengaged from the cylinder, leaving only index, middle, and ring fingers in contact.
    
\textit{(b) Tight 4-finger}: a four-finger grasp in which the contacts occupy a partial angular arc, producing a non-degenerate but unevenly distributed wrench-generation capacity.
    
\textit{(c) Spread 4-finger}: the Allegro Hand's natural four-finger antipodal grasp, with fingertips $\sim 90^\circ$ apart around the cylinder.

\textbf{Views per configuration (rows).}
    
\textit{First row}: wrench ellipsoid in the $(\tau_x,\tau_y,\tau_z)$ 3D-moment subspace, with principal semi-axes $(\sqrt{\lambda_4},\sqrt{\lambda_5},\sqrt{\lambda_6})$ of $\mathbf{M}$. The translucent outer surface represents the full moment-generation capacity; the opaque inner sphere of radius $\sqrt{\lambda_{\min}(\mathbf{M})}$ visualizes the worst-case wrench guarantee --- the maximum wrench magnitude the grasp can resist isotropically in any moment direction, i.e.\ the quadratic analog of the Ferrari--Canny enclosed-ball metric~\citep{219918}. The bold red double-arrow marks the $\lambda_{\min}(\mathbf{M})$ principal direction. 

\textit{Callout plots in the first row}: top-down schematic with the hand mesh tinted by the configuration's $\lambda_{\min}(\mathbf{M})$ color and the cylindrical object as a semi-transparent disk. Red dots mark the fingertip--object contact points; red arrowheads indicate inward contact normals projected into the tangent plane.

\textit{Second row}: full eigenvalue spectrum of $\mathbf{M}$, partitioned by the dashed gray line into the force block ($\lambda_1,\lambda_2,\lambda_3\approx\sqrt{n_\text{contact}}$, dominated by contact count rather than contact geometry) and the moment block ($\lambda_4,\lambda_5,\lambda_6$, which depend on contact distribution and carry the grasp-quality signal). $\lambda_{\min}(\mathbf{M})$ is highlighted in the configuration's color.

\textbf{Observation.} $\lambda_{\min}(\mathbf{M})$ increases by $5.3\times$ across the three configurations. The decrement from (b) to (a) corresponds to the structural transition from four-contact to three-contact grasping and is responsible for the largest single drop; the decrement from (c) to (b) is smaller and reflects only contact-geometry reorganization within the four-contact regime. The inscribed sphere in the moment ellipsoid contracts visibly as $\lambda_{\min}(\mathbf{M})$ decreases --- most pronounced in (a) --- directly illustrating the meaning of $\lambda_{\min}(\mathbf{M})$ as the squared radius of the largest isotropic wrench guarantee.

\subsection{Details of the Local Contact-Geometry Prior}
\label{app:mobility}

We expand the local contact-geometry analysis from
Sec.~\ref{subsec:contact_geometry}. Recall that the relative rolling mobility along a tangential direction is governed by the relative curvature
$\kappa_o+\kappa_f$ in that direction: zero relative curvature gives zero
mobility (no local rolling), while larger relative curvature increases mobility, localizes contact, and makes the motion more sensitive. Table~\ref{tab:mobility_summary} compares the relative curvature and rolling mobility of four fingertip morphologies along $X_c$ and $Y_c$ for a cylindrical object, where $K_o=\mathrm{diag}(\kappa_o,0)$ and $\kappa_o>0$. 

\begin{wrapfigure}{r}{0.35\textwidth}
    \centering
    \vspace{-0.15in}
    \includegraphics[width=0.35\textwidth]{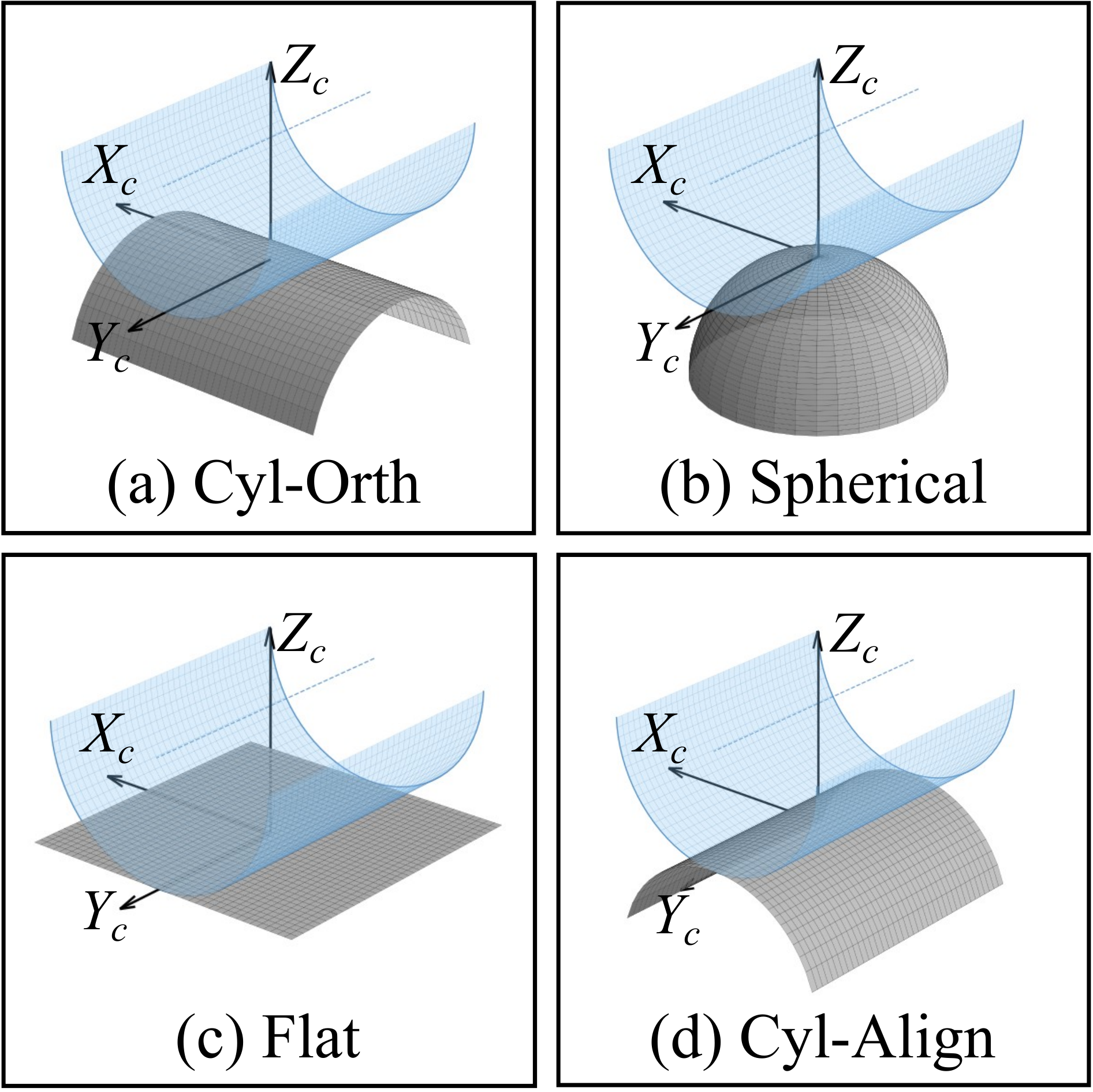}
    \vspace{-0.15in}
    \caption{Fingertip-object contact geometries. Gray is fingertip curvature and blue is object curvature.}
    \vspace{-0.25in}
    \label{fig:3d_shapes}
\end{wrapfigure}

\begin{table}[h]
\centering
\caption{Relative curvature and rolling mobility of four morphologies
on a cylindrical object.}
\label{tab:mobility_summary}
\begin{tabular}{lcccc}
\toprule
Fingertip & $X_c$ rel.\ curvature & $Y_c$ rel.\ curvature & $X_c$ mobility & $Y_c$ mobility \\
\midrule
Flat    & $\kappa_o$            & $0$        & moderate & zero \\
Spherical  & $\kappa_o+\kappa_f$   & $\kappa_f$ & high & nonzero \\
Aligned & $\kappa_o+\kappa_f$   & $0$        & high & zero \\
Orth    & $\kappa_o$            & $\kappa_f$ & moderate & nonzero \\
\bottomrule
\end{tabular}
\end{table}

\paragraph{Off-axis behavior ($Y_c$).}
Since the object is curved along $X_c$, all four morphologies retain rolling mobility along the task direction; their main difference appears along the off-axis direction $Y_c$. Off-axis relative curvature vanishes only when both surfaces are flat along $Y_c$, as with the flat and aligned fingertips. These morphologies therefore suppress local off-axis rolling passively. In contrast, the spherical and orthogonal fingertips introduce curvature along $Y_c$, allowing off-axis rolling and providing no directional preference. 

\paragraph{Task-aligned behavior ($X_c$).}
We next compare the flat and aligned fingertips, which share the same off-axis suppression but differ in task-aligned relative curvature: $\kappa_o$ for the flat fingertip and $\kappa_o+\kappa_f$ for the aligned fingertip. The larger relative curvature of the aligned fingertip keeps the contact point more localized on the fingertip during rolling. Under pure rolling, the fingertip arc length traversed by a task-aligned rolling angle $\theta$ is 
\begin{equation}
    s_f = \frac{\theta}{\kappa_o+\kappa_f},
    \label{eq:arc_relation}
\end{equation}
and thus
\begin{equation}
    s_f^{\mathrm{aligned}} = \frac{\theta}{\kappa_o+\kappa_f}
    \;<\;
    s_f^{\mathrm{flat}} = \frac{\theta}{\kappa_o}
    \quad (\kappa_f>0).
\end{equation}
For the same task-aligned rotation, the contact point migrates less on the aligned fingertip, allowing it to remain within the usable contact region longer. Thus, larger curvature along $X_c$ is preferable not for faster rolling, but for more localized and controlled contact.

\paragraph{Relation to compliant systems.}
Eq.~\eqref{eq:arc_relation} assumes a rigid point-contact model. In the real compliant system, a flat fingertip pressed against a cylindrical object forms an extended, poorly localized contact that can drift, while a curved fingertip produces a more localized and well-defined contact. Although this effect is not captured by the rigid model, both analyses favor the aligned cylindrical design. Consistently, Table~\ref{tab:morph_gen} shows that the aligned fingertip achieves the highest average success rate ($83\%$), outperforming the spherical ($53\%$), flat ($49\%$), and orthogonal ($43\%$) fingertips.

\addcontentsline{toc}{section}{Appendix}

\section{Policy Implementation Details}
\label{app:policy_implementation}
\subsection{Simulation Environment}
Policies are trained in IsaacGym Preview~4 with 8193 parallel environments on
a single RTX~3090. The simulator runs at $120~\mathrm{Hz}$, the policy is
queried at $20~\mathrm{Hz}$, and each episode lasts $400$ policy steps
($\approx 20~\mathrm{s}$). Domain-randomization ranges and PPO hyperparameters
are listed in Table~\ref{tab:app_hyperparams}.

\subsection{RL Formulation Details}
\textbf{Task.}
\label{subsubsec:hand_local_axis}
The goal is rotating an object about a hand-local task axis $\hat{\mathbf{a}}$ as in Sec.~\ref{subsec:modeling_frames}. The task axis $\hat{\mathbf{a}}$ in the world frame at step $t$ is
\begin{equation}
    \hat{\mathbf{n}}_{\mathrm{target}}(t)
    \;=\; s \,\cdot\, \mathbf{R}_{wh}(t)\,\hat{\mathbf{a}},
    \label{eq:rot_axis}
\end{equation}
where $\mathbf{R}_{wh}(t)$ maps vectors from hand frame $\mathcal{F}_h$ to world frame $\mathcal{F}_w$, and $s\in\{+1,-1\}$ sets the desired rotation direction. This hand-frame target decouples in-hand rotation from the arm's global state.

\begin{figure}[htbp]
\centering
\includegraphics[width=0.95\linewidth]{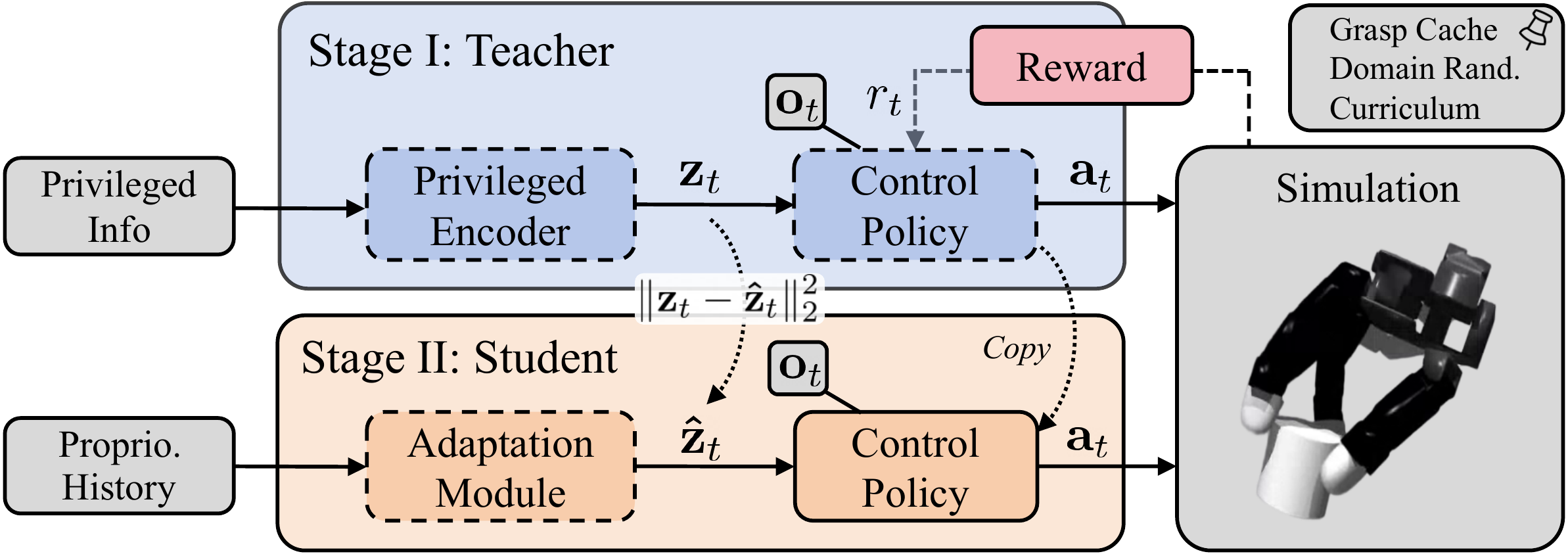}
\caption{\textbf{Training pipeline.} Dashed blocks are trainable components in each stage, and gray blocks indicate information or strategies associated with simulation. 
Each episode is initialized from the grasp cache with domain randomization under the current curriculum stage (top right). In Stage~I, the teacher is trained with PPO: a privileged encoder maps the privileged information to a latent $\mathbf{z}_t$; together with observation $\mathbf{o}_t$ from a single frame of joint positions and joint targets, the teacher control policy produces actions $\mathbf{a}_t$ under the reward $r_t$, including the grasp-quality shaping term. In Stage~II, an adaptation module regresses $\hat{\mathbf{z}}_t$ using only the proprioceptive history, supervised against the privileged latent $\mathbf{z}_t$; the student control policy then maps $(\mathbf{o}_t, \hat{\mathbf{z}}_t)$ to actions, requiring no privileged information at deployment. The simulation returns the latest joint positions and joint targets, which accumulate into the proprioceptive history and update the observation $\mathbf{o}_t$.}
\label{fig:method_overview}
\end{figure}

\textbf{RMA implementation details.} Fig.~\ref{fig:method_overview} summarizes the pipeline.   
In Stage~I, the teacher policy is conditioned on a single proprioceptive frame, consisting of 16 normalized joint positions and 16 joint targets, together with an 8D latent from a privileged encoder that takes privileged information in simulation, including object pose, scale, mass, friction coefficient, and center-of-mass offset. The teacher policy and privileged encoder are trained jointly with PPO. 

In Stage~II, the student uses the same actor--critic structure but replaces the privileged latent with an estimate from an adaptation module that uses only proprioceptive history. The module applies a 1D convolution followed by an MLP to a rolling buffer of the last $30$ proprioceptive frames. The student rollouts train the adaptation module by supervised regression to the frozen privileged latent, while the actor and critic are initialized from Stage~I and fine-tuned with PPO using the estimated latent. 

At deployment, the student policy transfers to the real world using only proprioceptive history, without privileged object parameters, tactile signals, visual observations, or object-state measurements. 

\begin{wraptable}{r}{0.5\textwidth}
\centering
\vspace{-0.28in}
  \caption{Domain randomization ranges}
  \label{tab:app_dr}
  \begin{tabular}{@{}lr@{}}
    \toprule
    Parameter & Value \\
    \midrule
    Object scale          & $\{0.7,\dots,0.8\}\pm\mathcal{U}[0.025]$ \\
    Object mass           & $\mathcal{U}[0.01,0.25]~\mathrm{kg}$ \\
    Object friction       & $\mathcal{U}[0.3,3.0]$ \\
    Object CoM offset     & $\mathcal{U}[-0.01,0.01]~\mathrm{m}$ \\
    PD P gain             & $\mathcal{U}[2.9,3.1]$ \\
    PD D gain             & $\mathcal{U}[0.09,0.11]$ \\
    External force        & every $0.05~\mathrm{s}$, decay $0.9$ \\
    \bottomrule
  \end{tabular}
  \vspace{-0.25in}
\end{wraptable}

\textbf{Domain randomization and disturbances.}
To prevent the teacher from overfitting to a nominal simulator and to give the student a nontrivial inference target, we randomize at each reset: object scale, object mass, friction, center-of-mass offset, and PD gains. External random forces are additionally applied to the object. The randomization ranges are given in Table~\ref{tab:app_dr}.

\textbf{Termination.}
An episode terminates when the object falls below a height threshold, when its world-frame displacement from the reset position exceeds a bound $\rho_{\mathrm{pos}}$, or when the episode-length cap is reached. The displacement bound acts as a loose workspace constraint that distinguishes genuine rolling, which is approximately stationary in position, from inadvertent ejection.

\textbf{Hand-orientation curriculum.} 
In hand-orientation curriculum, an episode succeeds if the object completes at least one full on-axis rotation. We track the success rate over a sliding window and smooth it with an exponential moving average $\widehat{\mathrm{SR}}_k$ using decay $\beta$. Stage $k$ advances when this smoothed success rate exceeds the target $\mathrm{SR}^\star_k$ after a minimum number of steps, or when the per-stage step cap is reached. Once stage $k=3$ is reached, the orientation bound stays at $180^\circ$ for the remainder of training.

\subsection{Reward Details}
\label{app:rewards}
\textbf{Task term.}
Let $\boldsymbol{\omega}_{\mathrm{obj}}(t)\in\mathbb{R}^3$ denote the object angular velocity in the  world frame, and let $\hat{\mathbf{n}}_{\mathrm{target}}(t)$ denote the desired task axis for rotation, also expressed in the world frame. We decompose $\boldsymbol{\omega}_{\mathrm{obj}}$ into components parallel and orthogonal to the target axis: 
\begin{equation}
    \boldsymbol{\omega}_{\parallel}(t)
    =
    \mathbf{P}_{\parallel}(t)\boldsymbol{\omega}_{\mathrm{obj}}(t),
    \qquad
    \boldsymbol{\omega}_{\perp}(t)
    =
    \mathbf{P}_{\perp}(t)\boldsymbol{\omega}_{\mathrm{obj}}(t),
\end{equation}
where the projection matrices are defined as
\begin{equation}
    \mathbf{P}_{\parallel}(t)
    =
    \hat{\mathbf{n}}_{\mathrm{target}}(t)
    \hat{\mathbf{n}}_{\mathrm{target}}(t)^\top,
    \qquad
    \mathbf{P}_{\perp}(t)
    =
    \mathbf{I}
    -
    \hat{\mathbf{n}}_{\mathrm{target}}(t)
    \hat{\mathbf{n}}_{\mathrm{target}}(t)^\top .
\end{equation}
The task term rewards angular velocity along the desired rotation direction while penalizing off-axis rotation:
\begin{equation}
    r_{\mathrm{task}}(t)
    =
    \alpha_{\parallel}
    \left[
    \max\left(
    0,
    \boldsymbol{\omega}_{\parallel}(t)^\top
    \hat{\mathbf{n}}_{\mathrm{target}}(t)
    \right)
    \right]^2
    -
    \alpha_{\perp}
    \left\|
    \boldsymbol{\omega}_{\perp}(t)
    \right\|^2 .
\end{equation}
Here, $\alpha_{\parallel}, \alpha_{\perp} > 0$ are weighting coefficients, and the $\max(\cdot)$ operator ensures that reward is assigned only to rotation in the desired direction.

\textbf{Grasp-quality term.}
The grasp-quality shaping term in Eq.~\eqref{eq:fc_reward} uses the minimum
eigenvalue of the hand-frame grasp Gramian $\mathbf{M}_h(t)=\mathbf{G}_h(t)\mathbf{G}_h(t)^\top$
as its argument. The function $\psi(\cdot)$ maps this raw grasp-quality value to a
bounded, monotonically increasing reward via an adaptive scale and a final clip:
\begin{equation}
    \psi\big(\lambda_{\min}(\mathbf{M}_h(t))\big)
    \;=\;
    \mathrm{clip}\Big(
        s_t\,\big(\lambda_{\min}(\mathbf{M}_h(t))-\epsilon\big),
        \;-r_{\max},\,r_{\max}
    \Big),
    \label{eq:psi_form}
\end{equation}
where $\epsilon$ is a small margin requiring the grasp to be conditioned beyond a
minimum level, and $r_{\max}$ bounds the term so that grasp quality is not
over-optimized at the expense of rotation progress. The factor $s_t$ adaptively
normalizes the term's magnitude through an exponential moving average (EMA) of its
recent unscaled values, keeping it on a comparable scale to the task reward across
training:
\begin{equation}
    s_t \;=\; \mathrm{clip}\!\left(\frac{m^{\star}}{10^{-6}+\bar m_t},\,0.1,\,20\right),
    \quad
    \bar m_t \;=\; \beta_{\mathrm{EMA}}\,\bar m_{t-1}
    \;+\;(1-\beta_{\mathrm{EMA}})\,\big|\lambda_{\min}(\mathbf{M}_h(t))-\epsilon\big|.
\end{equation}
A higher $\lambda_{\min}(\mathbf{M}_h)$ indicates a more balanced, well-conditioned
grasp and yields a larger reward, while values approaching degeneracy
($\lambda_{\min}\!\to\!0$) are penalized.
($\epsilon=10^{-4}$, $r_{\max}=0.5$, $m^{\star}=0.1$, $\beta_{\mathrm{EMA}}=0.97$).

\textbf{Regularization terms.}
A small set of regularization terms enforces safe and efficient execution,
\begin{equation}
    r_{\mathrm{reg}}(t)
    \;=\;
    -\lambda_{\mathrm{vel}}\|\mathbf{v}_{\mathrm{obj}}(t)\|^2
    \;-\;\lambda_{\mathrm{pose}}\|\mathbf{q}_{\mathrm{hand}}(t)-\mathbf{q}_{\mathrm{hand}}^{\mathrm{cache}}\|^2
    \;-\;\lambda_{\tau}\|\boldsymbol{\tau}(t)\|^2
    \;-\;\lambda_W\,W(t)
\end{equation}
where $\mathbf{v}_{\mathrm{obj}}$ is the object linear velocity, $\mathbf{q}_{\mathrm{hand}}^{\mathrm{cache}}$ is the cached initial hand joint configuration, $\boldsymbol{\tau}$ is the joint torque, and $W(t)$ is the instantaneous mechanical work. The velocity term discourages translation instead of rolling, consistent with the quasi-static assumption. The pose term keeps the policy near the cache-validated grasp, while the torque and work terms regularize control effort.

\textbf{Composite reward.}
The overall reward combines the three components above,
\begin{equation}
    r(t) \;=\; r_{\mathrm{task}}(t) \;+\; r_{\mathrm{gq}}(t) \;+\; r_{\mathrm{reg}}(t).
    \label{eq:reward_total}
\end{equation}
Among these, only $r_{\mathrm{gq}}$ is physics-derived in the sense of Sec.~\ref{sec:modeling}; it is the algorithmic expression of the grasp-quality prior.

\subsection{Training Configurations}

We provide supplementary information about the training details in the Table~\ref{tab:app_hyperparams}.

\begin{table}[htbp]
  \centering
  \caption{Architecture, PPO hyperparameters, and training budget for the two-stage RMA pipeline.}
  \label{tab:app_hyperparams}
  \begin{tabular}{@{}lr@{\hspace{2em}}lr@{}}
    \toprule
    \multicolumn{2}{c}{\textbf{Teacher (Stage~I)}} & \multicolumn{2}{c}{\textbf{Student (Stage~II)}} \\
    \cmidrule(r){1-2}\cmidrule(l){3-4}
    Encoder hidden   & $[256,128,8]$   & Adaptation module & TCN \\
    Policy hidden    & $[512,256,128]$ & Latent dim        & $8$ \\
    Activation       & ELU             & Policy hidden     & $[512,256,128]$ \\
    Learning rate    & $5\times10^{-4}$ & Activation       & ELU \\
    Clip ratio       & $0.2$           & Learning rate     & $3\times10^{-4}$ \\
    GAE $\lambda$    & $0.95$          & History window    & $30$ \\
    Discount         & $0.99$          & Optimizer         & Adam \\
    Entropy coef.    & $0.0$           & Num envs          & $20000$ \\
    Value loss coef. & $4$             & Env steps         & $1.5\times10^{9}$ \\
    Max grad norm    & $1.0$           & Mini epochs       & $1$ \\
    Optimizer        & Adam            & & \\
    Num envs         & $8193$          & & \\
    Env steps        & $1.5\times10^{9}$ & & \\
    Rollout steps    & $8$             & & \\
    Minibatch size   & $32{,}768$      & & \\
    Mini epochs      & $5$             & & \\
    \bottomrule
  \end{tabular}
\end{table}

\section{Fingertip Design and Fabrication}
\label{app:fingertip_design}
Each fingertip is fabricated as a single-piece silicone casting in a reusable multi-part mold. The mold consists of a base plate, a two-piece outer cavity, and an inner insert that defines the internal mounting geometry. All components are 3D-printed in PLA, lightly sanded along the parting surfaces, and coated with release agent before casting. The same mold base and insert are used across all fingertip designs; only the outer cavity halves are exchanged to realize different surface morphologies.

\begin{figure}[tbp]
    \centering
    \includegraphics[width=0.88\linewidth]{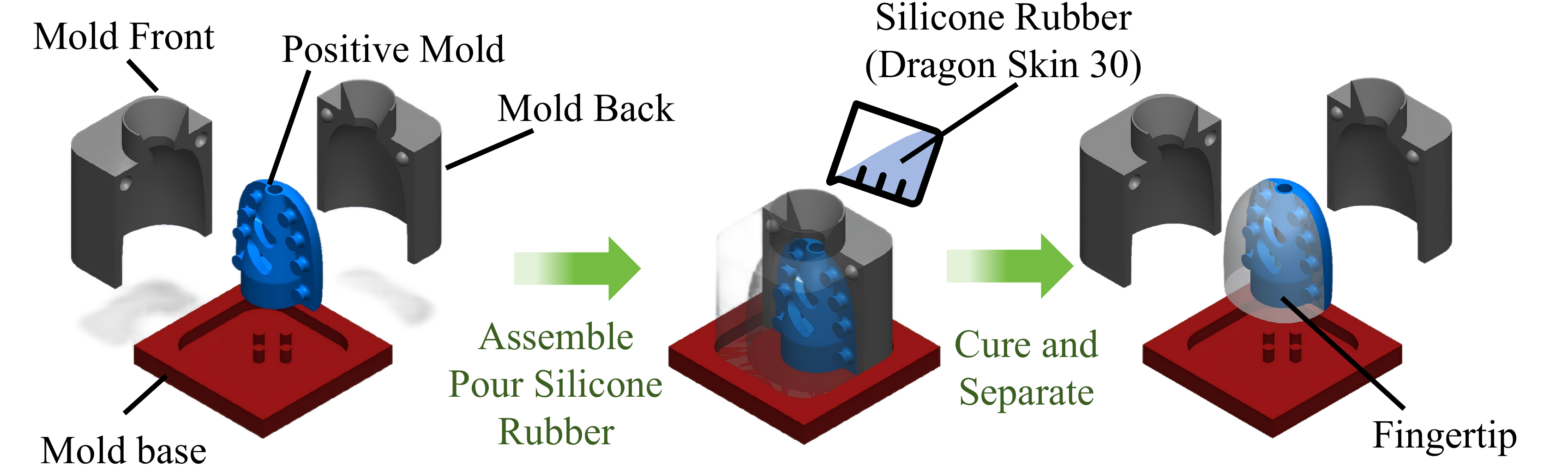}
    \caption{Silicone-casting workflow for fingertips. \textbf{(Left)} Exploded view of the multi-part mold. \textbf{(Center)} Assembled mold after silicone is poured around the insert. \textbf{(Right)}  Demolded fingertip after curing and separation, preserving the outer surface while integrating the mounting interface for attachment to the Allegro Hand's fingertip flange.}
    \label{fig:fabrication}
\end{figure}

\begin{figure}[tbp]
    \centering
    \includegraphics[width=0.8\linewidth]{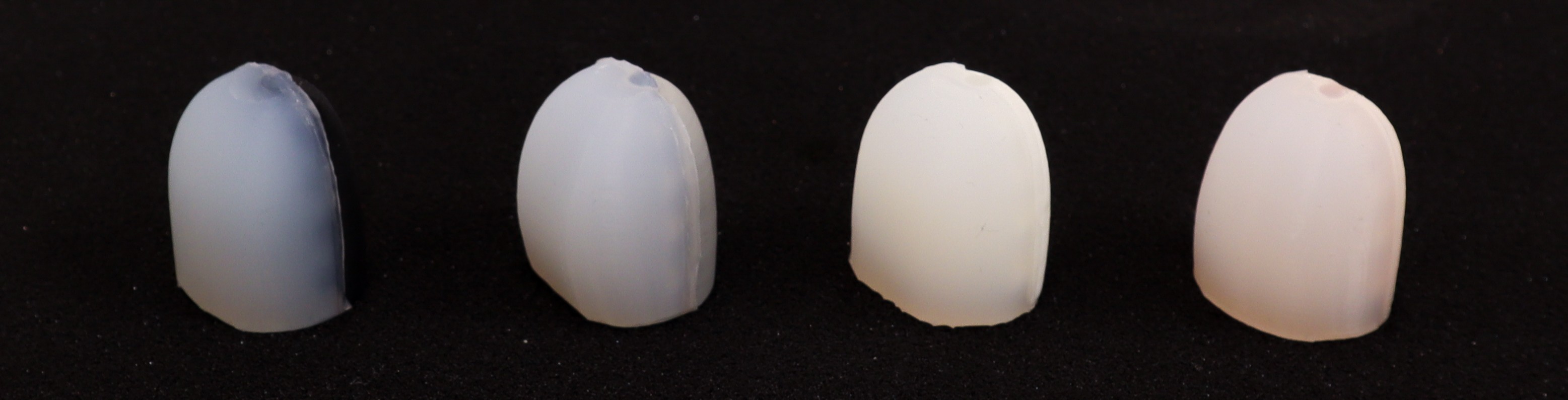}
    \caption{Fabricated fingertips. From left to right: Aligned cylindrical, orthogonal cylindrical, flat,  and spherical design. }
    \label{fig:fabricated_object}
\end{figure}

We cast the fingertips using Smooth-On Dragon Skin 30, a two-part platinum-cure silicone rubber with Shore 30A hardness. The two parts are hand-mixed at a $1{:}1$ mass ratio, vacuum-degassed for about five minutes, and poured around the inner insert into the assembled mold. The closed mold is then cured in room temperature for about $16$ hour. Brass heat-set threaded inserts are added to the inner mold after curing, leaving exposed threads for mounting to the Allegro Hand's distal joint. The fabrication process and resulting fingertips are shown in Fig.~\ref{fig:fabrication} and Fig.~\ref{fig:fabricated_object}. 

\begin{figure}[h!]
    \centering
    \includegraphics[width=0.8\linewidth]{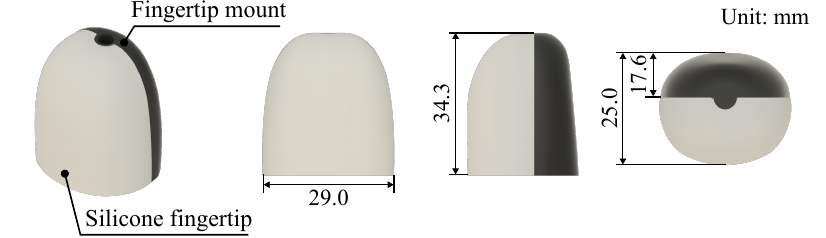}
    \caption{Dimensions of the aligned cylindrical fingertip designed for mounting on the distal joint of the Allegro Hand.}
    \label{fig:finger_dimension}
\end{figure}

The dimensions of the aligned cylindrical fingertip are shown in Fig.~\ref{fig:finger_dimension}. The other fingertips use the same overall dimensions, with only changing the contact surface geometry.

\section{Experimental Details}
\label{app:experiments}

\textbf{Hardware platform.} We use Allegro V4 four-fingered hand (16 active DoF) on a Franka Research~3 arm. The arm sets the global palm orientation before each rollout and stays fixed during the trial; a RealSense D435 logs an AprilTag~\cite{5979561} on the object for post-hoc pose measurement only, with no exteroceptive feedback to the policy.

\textbf{Task objects.} Real-world tests use three objects, including cylinder, cuboid, and ball. The cylinder is an empty aluminum can ($56~\mathrm{mm}$ diameter, $130~\mathrm{mm}$ height, $17~\mathrm{g}$). The cuboid is 3D-printed from PLA ($60~\mathrm{mm}\times60~\mathrm{mm}\times120~\mathrm{mm}$, $47~\mathrm{g}$), and the foam-rubber ball has a $50~\mathrm{mm}$ diameter and weighs $18~\mathrm{g}$. To increase contact friction, the cylinder and cuboid are wrapped with paper.

\textbf{Metrics.} For each rollout, we record the trajectory and object pose and compute all metrics post-hoc. \textbf{Rot} is the mean cumulative rotation about the task axis in turns. \textbf{Best} is the maximum absolute cumulative rotation across rollouts in a group. \textbf{TTT} is the mean episode duration, capped at $35~\mathrm{s}$ and independent of the rotation gate. \textbf{SR} is the gated success rate: a trial is successful only if it (i) maintains the grasp without an early drop or setup failure for at least $10~\mathrm{s}$ and (ii) achieves a rotation magnitude of at least $0.15$ turns. Holding the object without rotation is not counted as success.

\section{Extended Experimental Results}
\label{app:experiments_results}
\subsection{Simulation Results}

\begin{figure}[htbp]
    \centering
    \includegraphics[width=1.0\linewidth]{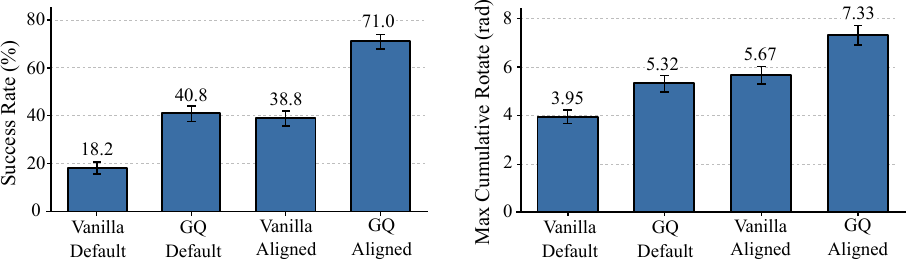}
    \caption{Simulation evaluation of the grasp-quality (GQ) prior with default and aligned fingertips. 
    \textbf{(Left)} Success rate, defined as completing at least one full turn without dropping or becoming stationary. \textbf{(Right)} Maximum cumulative rotation. Error bars are bootstrap $95\%$ CIs over $200$ episodes per configuration.}
    \label{fig:results_sim}
\end{figure}

In simulation, the performance trends are consistent with the real-world results (Fig.~\ref{fig:results_sim}). The baseline with the vanilla reward and default fingertip reaches an $18.2\%$ success rate, while the aligned-only and grasp-quality-only conditions reach $38.8\%$ and $40.8\%$, respectively. Combining the reward and fingertip designs further improves the success rate to $71.0\%$. The combined design also achieves a maximum cumulative rotation of $7.33~\mathrm{rad}$, compared with $3.95~\mathrm{rad}$ for the baseline. 
 
\subsection{Real-World Results}
\begin{figure}[tbhp]
  \centering
  \includegraphics[width=0.92\columnwidth]{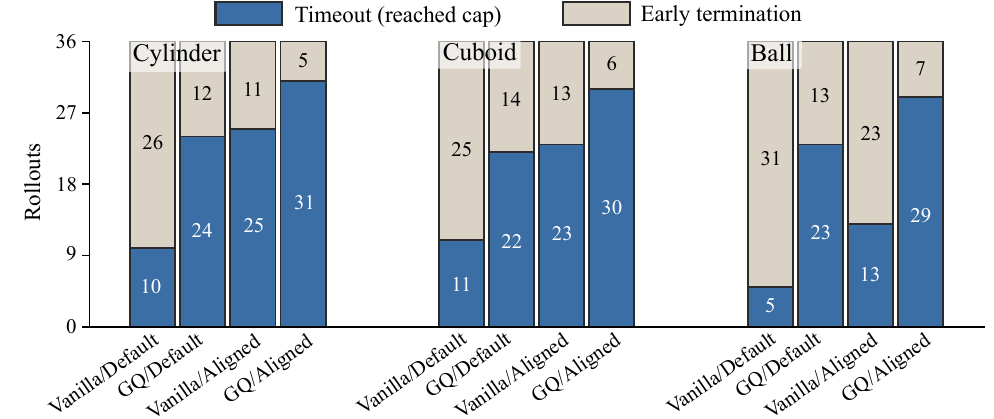}
  \caption{Termination breakdown for the grasp-quality prior with default and aligned fingertips across three objects.} 
  \label{fig:termination_grasp-quality}
\end{figure}

\begin{figure}[tbhp]
  \centering
  \includegraphics[width=0.92\columnwidth]{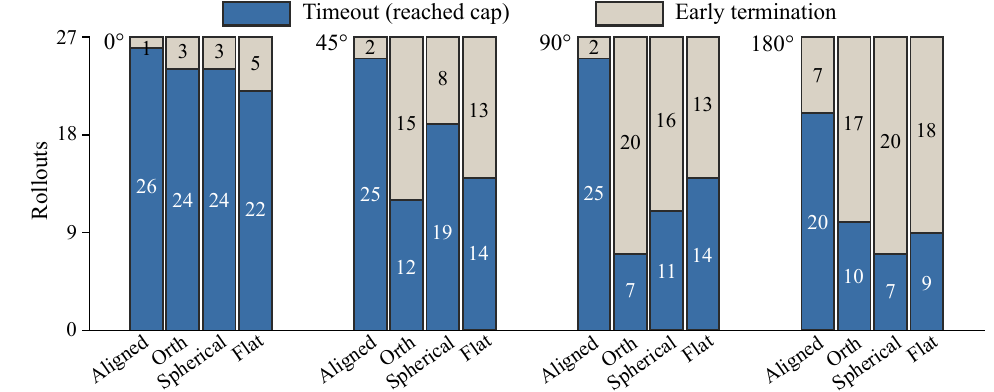}
  \caption{Termination breakdown for the fingertip morphology across four palm tilts.}
  \label{fig:termination_codesign}
\end{figure}

For real-world experiments, we provide the termination breakdown for both grasp-quality prior with default and aligned cylindrical fingertips (Fig.~\ref{fig:termination_grasp-quality}) and fingertip morphology design (Fig.~\ref{fig:termination_codesign}). 

Here, we also audit the detailed experimental results for the global grasp-quality prior (Table~\ref{tab:appendix_audit_codesign}) and local contact-geometry prior (Table~\ref{tab:appendix_morphology_per_tilt}), corresponding to Sec.~\ref{subsec:eval_grasp} and Sec.~\ref{subsec:eval_morph}, respectively.

\begin{table}[tbhp]
\centering
\caption{Audit table for global grasp-quality (GQ) prior evaluation using default and aligned fingertips, grouped by object. Bold marks the best value in each metric column per object.}
\label{tab:appendix_audit_codesign}
\begin{tabular}{ll cccc cc}
\toprule
& & \multicolumn{4}{c}{Metrics} & \multicolumn{2}{c}{Termination} \\
\cmidrule(lr){3-6}\cmidrule(lr){7-8}
Object & Config & Rot & Best & SR & TTT & Timeout & Early \\
\midrule
\multirow{4}{*}{Cylinder}
& A: Vanilla/Default & 1.61 & 3.55 & 28\% & 25.2 & 10 & 26 \\
& B: Vanilla/Aligned & 2.09 & 4.05 & 67\% & 26.3 & 24 & 12 \\
& C: GQ/Default      & 2.19 & 3.65 & 69\% & 28.8 & 25 & 11 \\
& D: GQ/Aligned      & \textbf{2.29} & 3.93 & \textbf{86\%} & \textbf{31.9} & \textbf{31} & 5 \\
\midrule
\multirow{4}{*}{Cuboid}
& A: Vanilla/Default & 0.27 & 1.01 & 31\% & 16.5 & 11 & 25 \\
& B: Vanilla/Aligned & 0.50 & 1.64 & 61\% & 27.0 & 22 & 14 \\
& C: GQ/Default      & \textbf{0.74} & 3.24 & 64\% & 25.9 & 23 & 13 \\
& D: GQ/Aligned      & 0.69 & 2.82 & \textbf{83\%} & \textbf{29.9} & \textbf{30} & 6 \\
\midrule
\multirow{4}{*}{Ball}
& A: Vanilla/Default & 0.60 & 2.97 & 14\% & 8.7  & 5  & 31 \\
& B: Vanilla/Aligned & 0.65 & 2.95 & 64\% & 25.8 & 23 & 13 \\
& C: GQ/Default      & \textbf{0.96} & 3.66 & 36\% & 16.2 & 13 & 23 \\
& D: GQ/Aligned      & 0.83 & \textbf{3.98} & \textbf{81\%} & \textbf{31.6} & \textbf{29} & 7 \\
\bottomrule
\end{tabular}
\end{table}

\begin{table}[tbhp]
\centering
\caption{Audit table for local contact-geometry prior evaluation. Rows are grouped by palm tilt; within each group, four fingertip morphologies are compared under rewards with the grasp-quality shaping. Bold marks the best value within each palm-tilt group.}
\label{tab:appendix_morphology_per_tilt}
\begin{tabular}{ll cc cc cc ccc}
\toprule
& & \multicolumn{2}{c}{Cylinder} & \multicolumn{2}{c}{Cuboid} & \multicolumn{2}{c}{Ball} & \multicolumn{3}{c}{Overall (27 runs)} \\
\cmidrule(lr){3-4}\cmidrule(lr){5-6}\cmidrule(lr){7-8}\cmidrule(lr){9-11}
Palm tilt & Fingertip & Rot & TTT & Rot & TTT & Rot & TTT & SR & Timeout & Early \\
\midrule
\multirow{4}{*}{0\textdegree}
& Aligned & \textbf{2.97} & 32.5 & \textbf{1.21} & \textbf{34.5} & 1.32 & \textbf{33.8} & \textbf{96\%} & \textbf{26} & 1 \\
& Orth    & 0.95 & 31.3 & 0.39 & 31.5 & \textbf{2.06} & 29.9 & 88\% & 24 & 3 \\
& Spherical  & 1.94 & \textbf{34.5} & 0.91 & 30.7 & 1.91 & 27.8 & 89\% & 24 & 3 \\
& Flat    & 1.94 & 29.9 & 0.91 & 34.2 & 1.12 & 22.6 & 83\% & 22 & 5 \\
\midrule
\multirow{4}{*}{45\textdegree}
& Aligned & \textbf{2.29} & \textbf{34.5} & \textbf{0.53} & \textbf{29.8} & \textbf{0.97} & \textbf{34.4} & \textbf{94\%} & \textbf{25} & 2 \\
& Orth    & 0.31 & 18.8 & -0.09 & 21.2 & 0.40 & 8.0 & 46\% & 12 & 15 \\
& Spherical  & 1.22 & 27.4 & 0.12 & 29.1 & 0.92 & 18.5 & 71\% & 19 & 8 \\
& Flat    & 0.30 & 10.6 & 0.29 & 29.6 & 0.72 & 13.1 & 51\% & 14 & 13 \\
\midrule
\multirow{4}{*}{90\textdegree}
& Aligned & \textbf{1.30} & \textbf{32.3} & \textbf{0.38} & \textbf{34.5} & \textbf{0.34} & \textbf{30.8} & \textbf{93\%} & \textbf{25} & 2 \\
& Orth    & 0.28 & 15.1 & 0.03 & 7.2 & 0.21 & 3.5 & 25\% & 7 & 20 \\
& Spherical  & 0.82 & 23.4 & -0.01 & 11.1 & 0.32 & 8.6 & 41\% & 11 & 16 \\
& Flat    & 0.64 & 23.1 & 0.04 & 24.5 & 0.17 & 4.9 & 50\% & 14 & 13 \\
\midrule
\multirow{4}{*}{180\textdegree}
& Aligned & 2.61 & 28.4 & \textbf{0.64} & \textbf{20.5} & 0.70 & \textbf{27.2} & \textbf{72\%} & \textbf{20} & 7 \\
& Orth    & 1.56 & 21.1 & 0.48 & 7.3 & \textbf{1.14} & 10.7 & 37\% & 10 & 17 \\
& Spherical  & 1.57 & 14.2 & 0.28 & 5.4 & 0.56 & 6.9 & 25\% & 7 & 20 \\
& Flat    & \textbf{2.88} & \textbf{29.7} & 0.36 & 3.6 & 0.26 & 1.9 & 34\% & 9 & 18 \\
\bottomrule
\end{tabular}
\end{table}

\subsection{Evaluation of Fingertip Morphology Under a Fixed Policy}
\label{app:isolate_fingertip_geometry}

We isolate the contribution of fingertip morphology by evaluating a single policy, which is trained on the default fingertip, across all four fingertip morphologies with only the physical fingertip swapped at test time (Table~\ref{tab:compare_only_hardware_overall}). This zero-shot transfer protocol holds the controller fixed, so performance differences can be attributed to local contact geometry. The aligned cylindrical fingertip achieves $2.57$ cumulative turns with the lowest variance ($\text{Std} = 0.10$) and runs productively up to the $35\,$s cap, whereas the flat, orthogonal, and spherical fingertips plateau below $1.7$ turns and terminate after only $40$--$48\%$ of the available window. This gap is consistent with the curvature analysis in Sec.~\ref{subsec:contact_geometry}: the aligned cylindrical fingertip passively supports task-aligned rolling and rejects out-of-plane disturbances, so a policy with no exposure to it still transfers cleanly. By contrast, the non-aligned geometries lose favorable contact interactions before sustaining continuous rotation. 

\begin{table}[tbhp]
\centering
\caption{Fingertip-swap comparison under a fixed policy. All morphologies are evaluated using the same policy trained with the default fingertip; only the physical fingertip is swapped at test time. Each row reports the average over three trials at a 180\textdegree{} palm tilt. \textbf{TTFT} is the mean time to the first full turn. 
Bold indicates the best value in each column.}
\label{tab:compare_only_hardware_overall}
\begin{tabular}{lcccccc}
\toprule
Fingertip & Rot $\uparrow$ & Std & Best $\uparrow$ & TTT $\uparrow$ & TTFT $\downarrow$ & SR $\uparrow$ \\
\midrule
Aligned & \textbf{2.57} & 0.10 & \textbf{2.66} & \textbf{34.5} & 14.1 & \textbf{99\%} \\
Orth & 1.30 & 0.36 & 1.64 & 14.1 & \textbf{10.4} & 40\% \\
Spherical & 1.34 & 0.13 & 1.45 & 15.2 & 11.1 & 44\% \\
Flat & 1.62 & 0.46 & 2.05 & 16.8 & 11.0 & 48\% \\
\bottomrule
\end{tabular}
\end{table}

\subsection{Experiment Time-lapse}
We show selected time-lapse sequences from the experiments used to evaluate the global grasp-quality prior and the local contact-geometry prior in Sec.~\ref{subsec:eval_grasp} and Sec.~\ref{subsec:eval_morph}. These examples illustrate representative in-hand rotation behaviors across reward shaping, fingertip designs, objects, and palm orientations. We also briefly discuss the failure modes observed during the experiments. 

\begin{figure}[tbhp]
  \centering
  \includegraphics[width=1.0\columnwidth]{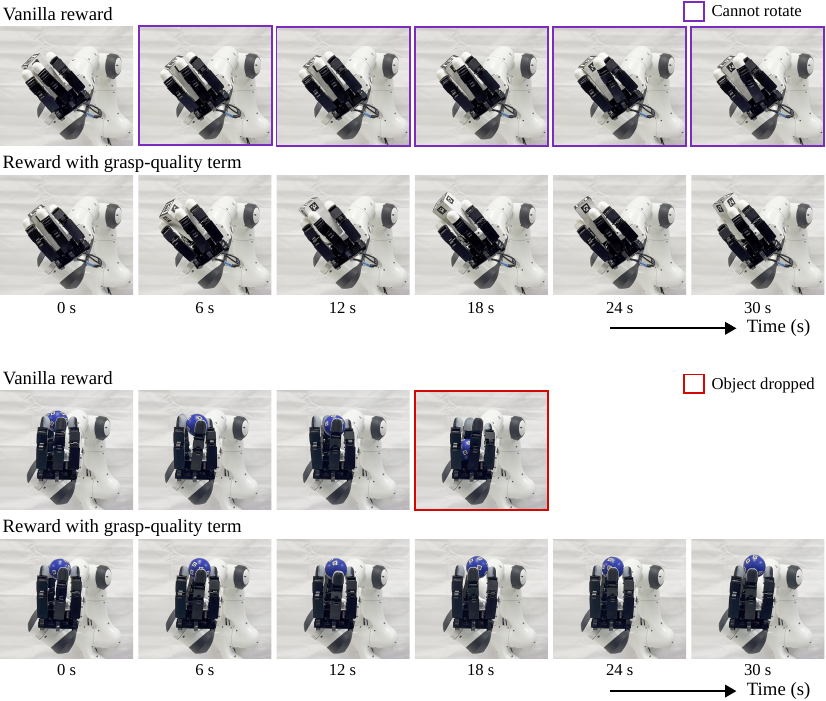}
  \caption{Time-lapse images of policy rollouts trained with and without grasp-quality reward shaping.} 
  \label{fig:time_lapse}
\end{figure}

Fig.~\ref{fig:time_lapse} presents two representative comparisons between policies trained with and without grasp-quality reward shaping. The top example shows cuboid rotation with the palm tilted by $45^\circ$ and the default fingertips. With the vanilla reward, the fingers fail to transition effectively across the cuboid edges and establish a stable grasp, leaving the object stationary throughout the rollout. Adding the grasp-quality term enables efficient rotation about the task axis. This comparison highlights the importance of maintaining grasp quality while pursuing the task objective. 

The bottom example shows ball rotation with the palm facing upward and using aligned cylindrical fingertips. The policy trained with the vanilla reward drops the ball midway through the rollout, while the policy trained with the grasp-quality term maintains a stable grasp and continues rotating the ball throughout the evaluation window. 

\begin{figure}[tbhp]
  \centering
  \includegraphics[width=1.0\columnwidth]{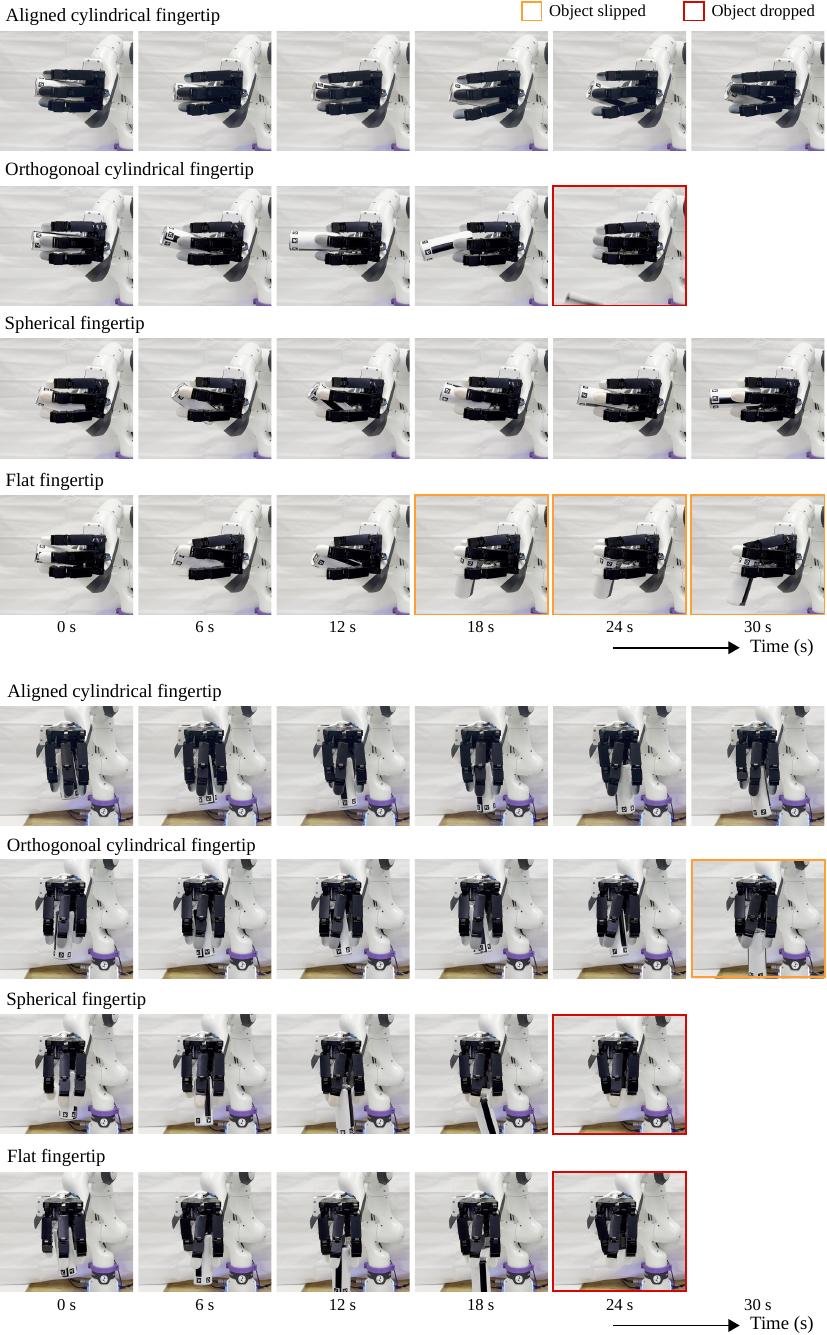}
  \caption{Time-lapse images of policy rollouts trained with different fingertip designs.} 
  \label{fig:time_lapse2}
\end{figure}

We further present two sets of comparisons across four fingertip designs in Fig.~\ref{fig:time_lapse2}. The top example rotates a cylinder object with a 90$^\circ$ palm tilt using policies only trained with the vanilla reward. In this configuration, the hand must resist gravity acting perpendicular to the task axis. The object slips and is dropped midway through the rollout with the orthogonal cylindrical and flat fingertips, empirically illustrating their limited robustness to off-axis disturbances. 

The bottom example shows cylinder rotation with a downward palm orientation using policies trained with the grasp-quality term, where gravity strongly destablizes the manipulation. With the orthogonal cylindrical fingertip, a sudden slip occurs between the 24~s and 30~s time stamps, causing the object to fall substantially. Both the spherical and flat fingertips fail to retain the object through the end of the evaluation while maintaining rotation.

\end{document}